\documentclass{article}

\usepackage{microtype}
\usepackage{graphicx}
\usepackage{subcaption}
\usepackage{booktabs} 
\usepackage{multirow}
\usepackage{hyperref}
\usepackage{adjustbox}

\usepackage[preprint]{icml2026}

\usepackage{amsmath}
\usepackage{amssymb}
\usepackage{mathtools}
\usepackage{amsthm}
\usepackage{csquotes}

\usepackage[capitalize,noabbrev]{cleveref}

\theoremstyle{plain}

\theoremstyle{definition}

\theoremstyle{remark}

\usepackage[textsize=tiny]{todonotes}

\icmltitlerunning{DualWeaver: Synergistic Feature Weaving Surrogates for Multivariate Forecasting with Uni-TSFM}

\begin{document}

\twocolumn[
  \icmltitle{DualWeaver: Synergistic Feature Weaving Surrogates for Multivariate Forecasting with Univariate Time Series Foundation Models}



  \icmlsetsymbol{equal}{*}

  \begin{icmlauthorlist}
    \icmlauthor{Jinpeng Li}{equal,yyy}
    \icmlauthor{Zhongyi Pei}{equal,yyy}
    \icmlauthor{Huaze Xue}{yyy}
    \icmlauthor{Bojian Zheng}{comp}
    \icmlauthor{Chen Wang}{yyy}
    \icmlauthor{Jianmin Wang}{yyy}
  \end{icmlauthorlist}

  \icmlaffiliation{yyy}{School of Software, BNRist, Tsinghua University.}
  \icmlaffiliation{comp}{Tencent}

  \icmlcorrespondingauthor{Zhongyi Pei}{peizhyi@tsinghua.edu.cn}

  \icmlkeywords{Multivariate Forecasting, Time Series Foundation Models, Adaptation, Feature Fusion}

  \vskip 0.3in
]



\printAffiliationsAndNotice{}  

\begin{abstract}
  Time-series foundation models (TSFMs) have achieved strong univariate forecasting through large-scale pre-training, yet effectively extending this success to multivariate forecasting remains challenging. 
  To address this, we propose DualWeaver, a novel framework that adapts univariate TSFMs (Uni-TSFMs) for multivariate forecasting by using a pair of learnable, structurally symmetric surrogate series. Generated by a shared auxiliary feature-fusion module that captures cross-variable dependencies, these surrogates are mapped to TSFM-compatible series via the forecasting objective. The symmetric structure enables parameter-free reconstruction of final predictions directly from the surrogates, without additional parametric decoding. A theoretically grounded regularization term is further introduced to enhance robustness against adaptation collapse. Extensive experiments on diverse real-world datasets show that DualWeaver outperforms state-of-the-art multivariate forecasters in both accuracy and stability. We release the code at \url{https://github.com/li-jinpeng/DualWeaver}.
\end{abstract}

\section{Introduction} \label{introduction}
Deep learning models have achieved remarkable success in diverse domains such as natural language processing (NLP) \cite{otter2020survey}, computer vision (CV) \cite{voulodimos2018deep}, and recommendation systems \cite{da2020recommendation}. Their ability to capture complex dependencies has also made them highly effective for time series analysis, surpassing traditional statistical methods \cite{wu2022timesnet, zeng2023transformers, liu2023koopa}. In particular, Transformer models have been successfully adapted for time series \cite{liu2023itransformer, nie2022time, wang_deep_2024}. Their attention mechanism enables selective focus on different parts of the input sequence, enabling flexible discovery of both temporal and cross-variable dependencies.

However, the potential of deep learning for time series forecasting is often limited by data sparsity. This challenge has been partially addressed by the emergence of large univariate time series foundation models (Uni-TSFMs) \cite{ansari_chronos_2024, TimerXL, liu2025sundial}. Pre-trained on massive datasets, these models exhibit strong zero-shot forecasting capabilities. They can be effectively adapted to downstream tasks by fine-tuning (updating a subset or all parameters), leveraging broad temporal knowledge while remaining robust with limited data \cite{ye2024survey, liang2024foundation}.

A key limitation of Uni-TSFMs, however, emerges in multivariate forecasting. In practice, variables from the same source are often inherently correlated, suggesting that modeling these dependencies should yield better performance than univariate (channel-independent) forecasting (we use channel interchangeably with variable, following previous work). Nevertheless, due to the architectural constraints for univariate temporal dynamics, Uni-TSFMs fundamentally overlook these crucial cross-variable dependencies \cite{marconi2025timeseriesfoundationmodels, feng2024only, benechehab_adapts_2025}.


Several recent studies have focused on modeling multivariate time series with TSFMs \cite{ekambaram2024tinytimemixersttms,liu_generalized_2024,benechehab_adapts_2025}. 
\textbf{Channel-independent fine-tuning} serves as a primary method for adapting TSFMs to specific target scenarios, whereas cross-variable dependencies are ignored. 
As a novel paradigm, AdaPTS \cite{benechehab_adapts_2025} introduces additional modules to capture cross-variable dependencies for a given vanilla or finetuned Uni-TSFM.
It employs \textbf{encoder-decoder adaptation}, in which the encoder first transforms original variables into a sequence of latent variables and the decoder generates final predictions based on these latent variables' predictions.
However, a key challenge remains: much of the adaptation capacity is consumed by reconstructing the real predictions from the latent variables, rather than by learning prediction-friendly cross-variable dependencies.

\begin{figure}[htbp]
    \centering
    \begin{subfigure}{0.41\columnwidth} 
        \includegraphics[width=\linewidth]{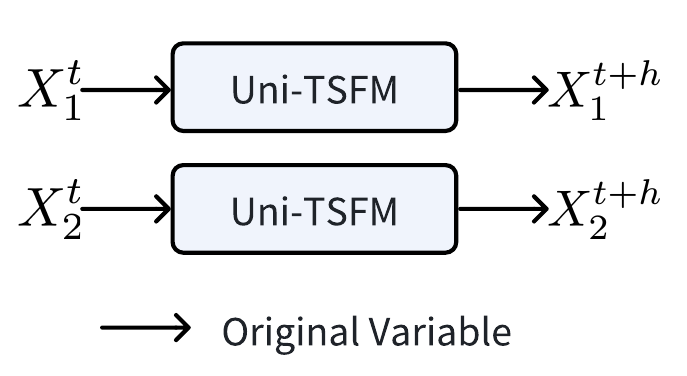}
        \caption{Channel-independent}
        \label{fig:a}
    \end{subfigure}
    \hspace{0.5mm}
    \begin{subfigure}{0.56\columnwidth} 
        \includegraphics[width=\linewidth]{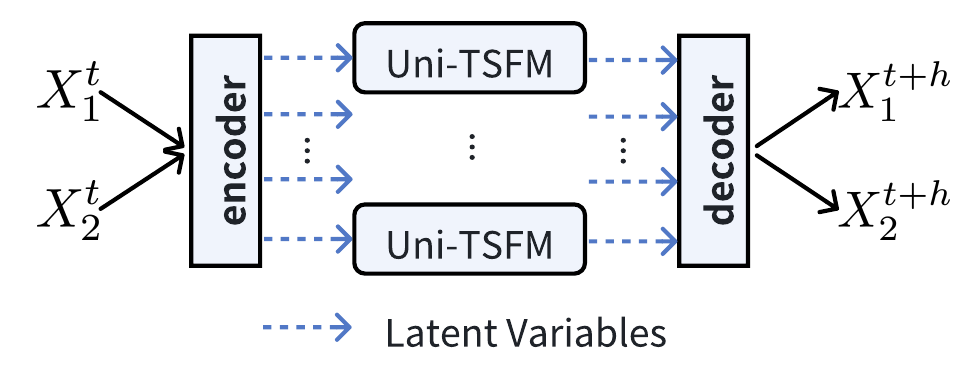}
        \caption{Encoder-Decoder}
        \label{fig:b}
    \end{subfigure}
    \begin{subfigure}{0.78\columnwidth} 
        \includegraphics[width=\linewidth]{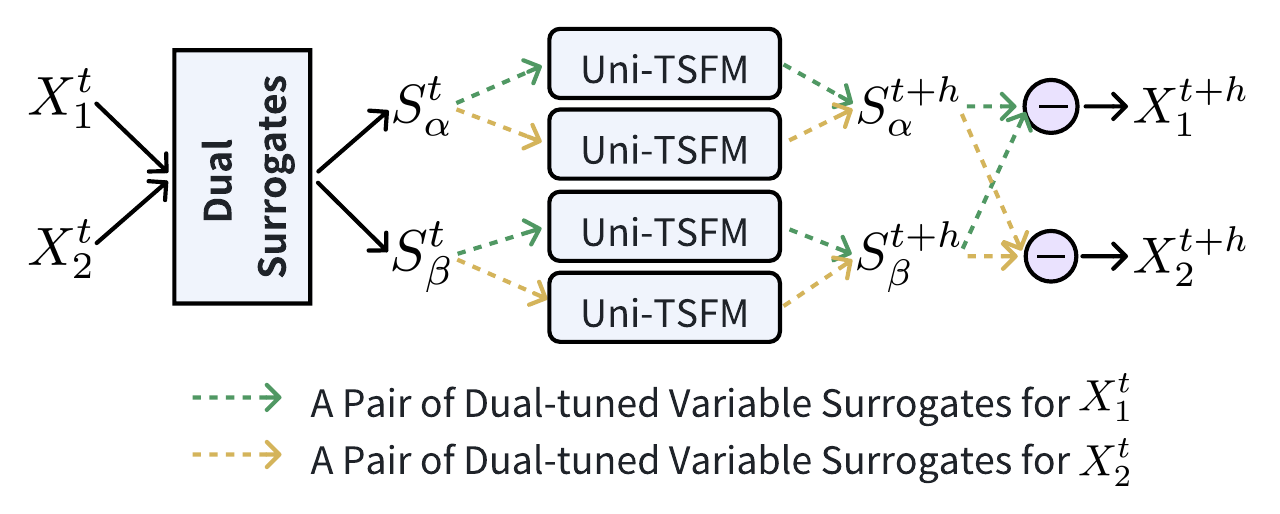}
        \caption{Dual-tuned Surrogates}
        \label{fig:c}
    \end{subfigure}
    \caption{The paradigms adapting Uni-TSFM for multivariate forecasting. (a) Channel-independent: each variable is processed separately by Uni-TSFMs, ignoring cross-variable dependencies. (b) Encoder-Decoder: an encoder maps original variables to latent representations, whose predictions are then decoded back; the decoding may mislead the adaptation. (c) Dual-tuned Surrogates (ours): a shared feature-fusion module generates dual surrogates that capture dependencies under distinct optimization directions, reducing overfitting and making adaptation more robust.}
    \label{fig:motivation}
\end{figure}

To address this, we propose the DualWeaver framework, which leverages a pair of \textbf{dual surrogates} to explicitly model cross-variable dependencies without relying on parametric reconstructions. Comparisons of the three paradigms are illustrated in Figure \ref{fig:motivation}.

A key design of DualWeaver is its shared feature-fusion module, which learns to fuse information across variables and outputs a pair of surrogate series. These surrogates, as more robust representations of the joint signal, simplify the problem for a Uni-TSFM, enabling accurate and stable reconstruction of the final multivariate forecasts.
The primary contributions of this work are:
\begin{itemize}
    \item We propose a novel adaptation paradigm for Uni-TSFMs, DualWeaver, that employs dual surrogates generated by a shared feature-fusion module to capture cross-variable dependencies and enable robust multivariate forecasting with given Uni-TSFMs.
    \item We provide a theoretical analysis of the error bound, which motivates the design of a novel regularization term. This term is explicitly formulated to stabilize training and enhance the robustness of the adapted model.
    \item Through extensive experiments on diverse benchmarks, we demonstrate that DualWeaver consistently outperforms state-of-the-art multivariate forecasters in both accuracy and robustness.
\end{itemize}

DualWeaver thus establishes a novel paradigm that effectively unlocks the potential of large pre-trained Uni-TSFMs for complex multivariate forecasting, directly addressing the core challenge of adapting powerful TSFMs with minimal cost while preserving their inherent strengths.

\section{Related Work}

\subsection{Time Series Foundation Models}

The field of time series forecasting is increasingly leveraging foundation models (FMs) to achieve more capable predictors \cite{ansari_chronos_2024,marconi2025timeseriesfoundationmodels,liang2024foundation}. These models are pre-trained on large datasets, which helps address the widespread issue of data sparsity.
Addressing the multivariate setting is particularly challenging for TSFMs, primarily due to increased data sparsity and the high complexity of model architecture. 

To the best of our knowledge, only a few TSFMs natively support multivariate analysis.
Moirai \cite{woo2024unifiedtraininguniversaltime} incurs a computational cost due to the simultaneous flattening of all channels, resulting in quadratic memory complexity with respect to the number of channels.
GTT \cite{feng2024only} employs shared-weight attention mechanisms to capture both temporal and cross-variable dependencies.
Chronos-2 \cite{chronos2} achieves unified modeling of both univariate and multivariate time series through a general-purpose architecture that sequentially applies attention across both the temporal and variate dimensions.

Although these TSFMs can achieve good performance on some multivariate forecasting benchmarks, they are typically designed to rely on a specific architecture to handle the complexity of multivariate time series, which limits their scalability to dynamic patterns in real-world multivariate scenarios.
Simultaneously, the potential of Uni-TSFMs for multivariate forecasting remains underexplored. Developing extensible frameworks to effectively harness the Uni-TSFMs' powerful univariate forecasting capabilities for multivariate tasks represents a promising and important research direction.

\subsection{Multivariate Forecasting Adapters}

Fine-tuning is the predominant approach for adapting TSFMs to downstream tasks. However, when applied to Uni-TSFMs for multivariate forecasting, standard fine-tuning fails to capture cross-variable dependencies essential for accurate predictions.
To bridge this gap, recent studies have introduced specialized adaptation mechanisms. 
For instance, TTM \cite{ekambaram2024tinytimemixersttms} employs channel mixing within a fine-tuned decoder head, leaving the pre-trained backbone restricted to channel-independent learning.
Gen-P-Tuning \cite{liu_generalized_2024} utilizes summary \enquote{prompts} as prefix sequences, yet this shallow adaptation is inherently constrained by the frozen model’s attention mechanism and a fixed prompt space.
AdaPTS \cite{benechehab_adapts_2025} wraps a Uni-TSFM in a trainable encoder-decoder structure to model variable interactions. However, its ablation study shows a decoder-only variant matches the full model's performance, suggesting the encoder fails to fully distill cross-variable dependencies.

In contrast, DualWeaver proactively transforms the multivariate into a pair of learnable, highly predictable surrogates via a shared feature-fusion module. Unlike the late-stage mixing of TTM or the prefix-based prompting of Gen-P-Tuning, our framework integrates deep cross-variable dependencies before foundation model processing. Crucially, DualWeaver enables non-parametric reconstruction, focusing the entire adaptation on leveraging variable dependencies instead of being distracted by a parametric decoder.

\section{DualWeaver Framework}

\subsection{Overview}

\begin{figure*}[htbp]
  \centering
  \includegraphics[width=\linewidth]{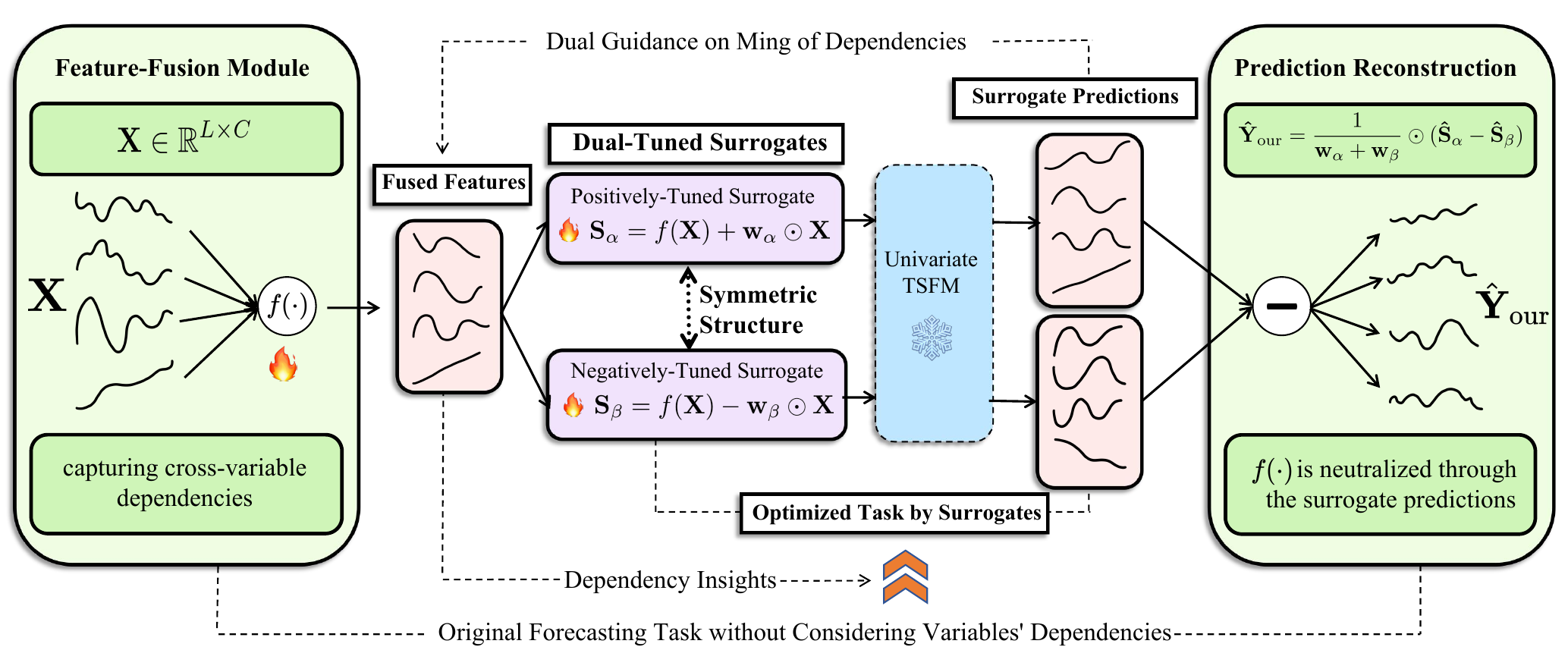}
  \caption{The overview of the DualWeaver framework. A shared feature-fusion module first extracts cross-variable dependencies to generate a pair of dual-tuned surrogates ($\mathbf{S}_\alpha$, $\mathbf{S}_\beta$). These surrogates are independently processed by a frozen Uni-TSFM, after which the final multivariate predictions are derived via non-parametric reconstruction to neutralize the shared fusion part.}
  \label{overview}
\end{figure*}

The proposed DualWeaver framework for adapting Uni-TSFMs for multivariate forecasting is presented in Figure \ref{overview}.
We denote the input multivariate time series data by $\mathbf{X}\in \mathbb R^{L\times C}$, where $L$ is the lookback window size in forecasting, and $C$ is the number of variables. The prediction of multivariate time series data is denoted by $\mathbf{\hat{Y}}\in \mathbb R^{H\times C}$, where $H$ is the future steps for prediction, and $\mathbf{Y}\in \mathbb R^{H\times C}$ denotes the ground truth. 
Specifically, we use $\mathbf{\hat{Y}}_\text{ori}$ to indicate the original prediction of a given TSFM $\mathcal M$ and $\mathbf{\hat{Y}}_\textbf{{our}}$ to denote the prediction of our work. 
Our goal is to reduce the generalization error of a given TSFM $\mathcal M$ on multivariate test sets while keeping $\mathcal M$ frozen.

DualWeaver captures cross-variable dependencies via a length-agnostic feature-fusion module $f(\cdot): \mathbb{R}^{T \times C} \rightarrow \mathbb{R}^{T \times C}$, where $T$ denotes an arbitrary sequence length (e.g., the lookback window $L$ or the ground truth horizon $H$). While $f(\cdot)$ preserves the dimensionality of the input sequence, it transforms the original features into a fused representation space.


We utilize the feature-fusion module to construct a pair of dual-tuned surrogates, a positively-tuned surrogate and a negatively-tuned surrogate, each of which is a specific instance of the proposed surrogate's hypothesis space, i.e., $\mathbf {S} = \mathbf{\alpha}_1   f(\mathbf {X}) + \mathbf{\alpha}_2 \mathbf{X}$.
For the positively-tuned surrogate, we define it by
\begin{equation} \label{alpha}
    \mathbf{S}_\alpha  = f(\mathbf X)+ \mathbf{w_\alpha} \odot \mathbf X
\end{equation}
where $\mathbf{w_\alpha} \in \mathbb{R}^C$ are linear channel-wise weights.
Similarly, we define the negatively-tuned surrogate by
\begin{equation} \label{beta}
    \mathbf{S}_\beta  = f(\mathbf{X})- \mathbf{w_\beta} \odot \mathbf{X}
\end{equation}
where $\mathbf w_\beta \in \mathbb{R}^C$ are also linear channel-wise weights.
The surrogates are viewed as the input and ground truth of multivariate forecasting. During adaptation, $f(\cdot)$, $\mathbf w_\alpha$, and $\mathbf w_\beta$ are optimized, while the given TSFM $\mathcal M$ 
(vanilla or fine-tuned) 
remains frozen.

The dual-tuned surrogates are designed to capture complementary aspects of cross-variable dependencies. From the viewpoint of function approximation, the dual surrogates define two distinct optimization directions in the feature space. A geometric interpretation is provided in Appendix \ref{geometric}.
The final prediction is derived via non-parametric reconstruction:
\begin{equation} \label{equal_y_our}
    \mathbf{ \hat{Y}_\text{our}}=\frac{1}{\mathbf w_\alpha+\mathbf w_\beta} \odot(\mathbf{\hat{S}_\alpha} - \mathbf{\hat{S}_\beta})
\end{equation}
where $\mathbf{ \hat{S}}$ denotes the prediction of the surrogates by $\mathcal M$.
Numerical stability is inherently guaranteed as the regularization term $\Omega$ (Eq. \ref{bound}) approaches infinity if $(\mathbf w_\alpha+\mathbf w_\beta) \rightarrow 0$, naturally preventing the denominator from vanishing during optimization.
Facilitating non-parametric reconstruction directly enables DualWeaver to learn cross-variable dependencies via its shared feature-fusion module robustly. Unlike encoder-decoder methods such as AdaPTS \cite{benechehab_adapts_2025}, which consume adaptation capacity to restore original variables, our framework prioritizes an efficient representation of these dependencies. This mechanism further mitigates overfitting by introducing distinct optimization directions, as evidenced by training stability in Appendix \ref{Training Stability From Dual-Surrogate}.

\subsection{Theoretical Analysis} \label{theory}

In practice, multiple variables from the same source often exhibit inherent correlations, which underpin the argument that channel-dependent forecasting should outperform channel-independent forecasting \cite{wang2024timexer, iTransformers}.
In this section, we propose a condition for this argument by analyzing the theoretical error bound of DualWeaver compared with univariate forecasting.
According to Eq. \ref{equal_y_our}, the ground truths for the dual-tuned surrogates are,


\begin{equation}
    \mathbf{\tilde{S}}_\alpha=f(\mathbf Y)+ \mathbf w_\alpha \odot \mathbf Y 
\end{equation}
\begin{equation}
    \mathbf{\tilde{S}}_\beta=f(\mathbf{Y})- \mathbf{w}_\beta \odot \mathbf Y
\end{equation}

Then we can define the difference between $\mathbf{\hat{Y}}_\text{our}$ and the ground truth on the $i$-th channel:
\begin{align}
   \mathbf {\hat{Y}}^i_\text{our} - \mathbf Y^i &= \frac{\mathbf {\hat{S}}_\alpha^i - \mathbf {\hat{S}}_\beta^i}{\mathbf w_\alpha^i+\mathbf w_\beta^i} - \frac{\mathbf{\tilde{S}}_\alpha^i - \mathbf{\tilde{S}}_\beta^i}{\mathbf w_\alpha^i+\mathbf w_\beta^i} \\
   &= \frac{1}{\mathbf w_\alpha^i+\mathbf w_\beta^i}[(\mathbf {\hat{S}}_\alpha^i - \mathbf{\tilde{S}}_\alpha^i) - (\mathbf {\hat{S}}_\beta^i - \mathbf{\tilde{S}}_\beta^i)]
\end{align}
The MSE of the DualWeaver framework on the $i$-th channel can be defined as follows:
\begin{align}
    \mathbf{E}_\text{our}^i &= \mathbb{E}[\left|\mathbf {\hat{Y}}_\text{our}^i - \mathbf Y^i\right|^2] \\
    &= \mathbb{E}[(\frac{1}{\mathbf w_\alpha^i+\mathbf w_\beta^i}[(\mathbf {\hat{S}}_\alpha^i - \mathbf{\tilde{S}}_\alpha^i) - (\mathbf {\hat{S}}_\beta^i - \mathbf{\tilde{S}}_\beta^i)])^2] \\
    &\le \frac{1}{(\mathbf w_\alpha^i+\mathbf w_\beta^i)^2}\mathbb{E}[(\left|\mathbf {\hat{S}}_\alpha^i - \mathbf{\tilde{S}}_\alpha^i\right| + \left|\mathbf {\hat{S}}_\beta^i - \mathbf{\tilde{S}}_\beta^i\right|)^2] \\
    &\le \frac{2}{(\mathbf w_\alpha^i+\mathbf w_\beta^i)^2}\mathbb{E}[(\mathbf {\hat{S}}_\alpha^i - \mathbf{\tilde{S}}_\alpha^i)^2 + (\mathbf{\hat{S}}_\beta^i - \mathbf{\tilde{S}}_\beta^i)^2]
\end{align}
where $\mathbb{E}[(\mathbf {\hat{S}}_\alpha^i - \mathbf{\tilde{S}}_\alpha^i)^2]$ and $\mathbb{E}[(\mathbf {\hat{S}}_\beta^i - \mathbf{\tilde{S}}_\beta^i)^2]$ are, respectively, the MSE of the positively-tuned surrogate $\mathbf{S}_\alpha$ and the negatively-tuned surrogate $\mathbf{S}_\beta$ on $i$-th channel.
Therefore, we define the theoretical error bound for the $i$-th channel as:
\begin{equation} \label{bound}
  \Omega_i = \frac{2}{(\mathbf w_\alpha^i+\mathbf w_\beta^i)^2}(\mathbf E_\alpha^i + \mathbf E_\beta^i)
\end{equation}
where $\mathbf E_\alpha^i$ and $\mathbf E_\beta^i$ denote the MSE of the surrogates in DualWeaver on the $i$-th channel. Then the sufficient condition for DualWeaver to have a lower error bound than that of the original channel-independent forecasting is as follows:
\begin{align} \label{condition}
\Omega_i &\le \mathbf E_{\text{ori}}^i, \quad 
i \in \{1, 2, ..., C\}
\end{align}
where $C$ denotes the number of input channel and $\mathbf E_\text{ori}^i$ denotes the MSE of the original channel-independent forecasting on the $i$-th channel.
As an easily achievable boundary case, let $\mathbf E_\alpha$ and $\mathbf E_\beta$ be equal to $\mathbf E_{\text{ori}}$, and $\mathbf w_\alpha^i+\mathbf w_\beta^i=2$ for $i \in \{1, 2, ..., $C$\}$, the condition will be satisfied. 
Incorporating this lower error bound condition of Eq. \ref{condition} as a regularization term ensures adaptation robustness, effectively averting the catastrophic divergence and exponential error growth empirically observed in Section \ref{Robustness from Error Bound Condition}.

\subsection{Multivariate Feature-Fusion Module} \label{weaver}

We define the multivariate feature-fusion module by $f(\cdot): \mathbb{R}^{T \times C} \rightarrow \mathbb{R}^{T \times C}$, which transforms the original multivariate into a fused feature space. For instance, we employ a multi-layer perceptron (MLP) architecture to capture cross-variable dependencies at each timestamp. Specifically, the MLP-based feature-fusion module processes the multivariate input at each time step $t$ as follows:
\begin{equation}
f(\mathbf{X})_{\text{MLP}}^{t} = \mathbf{W}_2 \times \sigma \left( \mathbf{W}_1 \times \mathbf{X}^{t} + \mathbf{b}_1 \right) + \mathbf{b}_2
\end{equation}
where ${\mathbf{W}_1} \in \mathbb{R}^{D \times C}$ and ${\mathbf{W}_2} \in \mathbb{R}^{C \times D}$ are weights, $D$ denotes the dimension of hidden layer, $\mathbf{b}_1 \in \mathbb{R}^{D}$ and $\mathbf{b}_2 \in \mathbb{R}^{C}$ are biases, and $\sigma$ is the activation function. 
It learns nonlinear timestamp-specific relationships across variables. 

Prior works have shown that diverse multivariate dependencies benefit from distinct modeling strategies \cite{shao_exploring_2025}.
Accordingly, DualWeaver is purposefully designed as an extensible framework to accommodate such variability. This extensibility is fundamentally rooted in the architectural substitutability of DualWeaver, which allows the feature-fusion module $f(\cdot)$ to be seamlessly replaced by alternative multivariate backbones, such as CNN-, GNN-, or Transformer-based models. To further demonstrate this architectural versatility, we provide additional evaluations of an alternative module design, a CNN-based implementation, in Appendix \ref{Extensibility}.

\subsection{Adaptation Process}

With the dual surrogates, the forecasting task is transformed from the original feature space into the surrogate space,
as shown in Eq. \ref{alpha} and \ref{beta}, capturing variables' dependencies through the feature-fusion module $f(\cdot)$. By initializing $f(\cdot)$ to zero, the framework ensures that the optimization process starts directly from the foundation model's original predictive state. This stability-preserving initialization allows the model to refine representations incrementally without noticably distorting the original series, thereby preserving the foundation model's inherent temporal stability and effectively shielding it from overfitting.

During the adaptation, the primary loss functions are the MSE of the predictions on the dual surrogates: 
\begin{equation}
\mathcal L_\alpha = \frac{1}{C} \sum_{i=1}^{C} \mathcal L_\alpha^i=\frac{1}{C} \sum_{i=1}^{C}\text{MSE}(\mathbf {\hat{S}}_\alpha^i, \mathbf {\tilde{S}}_\alpha^i) 
\end{equation}
\begin{equation}
\mathcal L_\beta = \frac{1}{C} \sum_{i=1}^{C} \mathcal L_\beta^i=\frac{1}{C} \sum_{i=1}^{C}\text{MSE}(\mathbf {\hat{S}}_\beta^i, \mathbf {\tilde{S}}_\beta^i) 
\end{equation}
where $\mathbf {\hat{S}}_\alpha^i=\mathcal M(\mathbf S_\alpha^i)$ and $\mathbf {\hat{S}}_\beta^i=\mathcal M(\mathbf S_\beta^i)$ denote the predictions of TSFM $\mathcal M$ on the $i$-th channel in surrogate space.
Only the feature-fusion module $f(\cdot)$ and the weights of linear combination in the surrogates, $\mathbf{w}_\alpha$ and $\mathbf{w}_\beta$, are optimized in the adaptation process.
As demonstrated in Section \ref{theory}, a lower error bound condition is also used as a regularization term in the optimization:
\begin{equation} \label{regularization}
\mathcal L_\text{bound} = \frac{1}{C} \sum_{i=1}^{C}\text{MAX}(\Omega_i, \text{MSE}(\mathbf {\hat{Y}}_\text{ori}^i,\mathbf Y^i))
\end{equation}

The total optimization objective is formulated as follows:
\begin{equation}
\mathcal L_\text{total} = \frac{\mathcal L_\alpha + \mathcal L_\beta}{2} + \lambda \mathcal L_\text{bound}
\end{equation}
where $\lambda$ is a hyperparameter used to regulate the influence of the theoretical error bound. To minimize the complexity of hyperparameter tuning and present the universality of the model, we adopt $\lambda=1$ as a natural weight, thus assigning equal priority to empirical loss minimization and maintaining the theoretical safety boundary.
Once the adaptation converges, a robust prediction of the test set of the original multivariate time series can be derived through Eq. \ref{equal_y_our}.

\section{Experiments}

In this section, we conduct a comprehensive evaluation of the proposed DualWeaver's performance across multiple publicly available datasets. The empirical results demonstrate that DualWeaver consistently achieves state-of-the-art performance across all evaluated datasets (Section \ref{Forecasting Performance}). Furthermore, DualWeaver exhibits enhanced robustness against the injection of irrelevant noisy sequences (Section \ref{Robustness to Irrelevant Dimensions}) and demonstrates superior computational efficiency (Section \ref{Computational Efficiency}). Finally, we present an ablation study to validate our framework's design, specifically highlighting the robustness afforded by our theoretical error bound condition (Section \ref{Robustness from Error Bound Condition}) and its exceptional scalability in high-dimensional settings (Section \ref{Dimensional Scalability}).

\subsection{Setup}
We evaluate the performance of DualWeaver in multivariate forecasting, leveraging two state-of-the-art Uni-TSFMs: Sundial \cite{liu2025sundial} and TimerXL \cite{TimerXL}. Benchmarks are conducted on core datasets, including the ETT (4 subsets) and Weather, and are supplemented by extended analyses on additional datasets, such as Electricity \cite{zhou2021informer, wu2021autoformer}. Our evaluation encompasses a comprehensive suite of baselines: (a) zero-shot and fully fine-tuned (SFT) configurations of the original Uni-TSFMs; (b) the state-of-the-art encoder-decoder method AdaPTS; and (c) leading native multivariate forecasters, including TimePro \cite{TimePro}, SimpleTM \cite{simpletm}, TimeMixer \cite{TimeMixer}, and iTransformer \cite{iTransformers}. All experiments are conducted on 8 NVIDIA H20 GPUs. In line with previous works, we employ the mean squared error (MSE) and the mean absolute error (MAE) as evaluation metrics. Detailed experimental configurations are provided in Appendix \ref{Implementation Details}. 

\subsection{Main Results}
\label{main results section}

\begin{table*}[htbp]
\begin{center}
\caption{Multivariate forecasting results on public datasets with forecasting horizons $\in \{96, 192, 336, 720\}$. Averaged results are reported here, and full results are provided in Table \ref{detail public dataset results} with the best in \textcolor{red}{\textbf{bold}} and the second \textcolor{blue}{\underline{underlined}}. 
$1^{st}$ Count represents the number of wins achieved by a method under all prediction lengths and datasets. 
Results for TimePro, SimpleTM, TimeMixer, and iTransformers are reported in \cite{TimePro} and \cite{simpletm}. Extended evaluations on \textbf{additional datasets} are provided in Appendix \ref{Evaluation on More Datasets}.}

\resizebox{\linewidth}{!}{
\begin{tabular}{c|c|cc|cc|cc|cc|cc|cc|cc|cc}

\toprule

\multicolumn{2}{c|}{\multirow{2}{*}{\textbf{Methods}}} & 
\multicolumn{2}{c}{\textbf{DualWeaver}} & 
\multicolumn{2}{c}{\textbf{AdaPTS}} & 
\multicolumn{2}{c}{\multirow{2}{*}{\textbf{SFT}}} & 
\multicolumn{2}{c}{\multirow{2}{*}{\textbf{Vanilla}}} & 
\multicolumn{2}{c}{\textbf{TimePro}} & 
\multicolumn{2}{c}{\textbf{SimpleTM}} & 
\multicolumn{2}{c}{\textbf{TimeMixer}} & 
\multicolumn{2}{c}{\textbf{iTrans.}} \\
\multicolumn{2}{c|}{} & 
\multicolumn{2}{c}{\textbf{(Ours)}} & 
\multicolumn{2}{c}{(2025)} & 
\multicolumn{2}{c}{} &  
\multicolumn{2}{c}{} &  
\multicolumn{2}{c}{(2025)} & 
\multicolumn{2}{c}{(2025)} & 
\multicolumn{2}{c}{(2024)} & 
\multicolumn{2}{c}{(2024c)} \\ 

\cmidrule(lr){1-2} \cmidrule(lr){3-4} \cmidrule(lr){5-6} \cmidrule(lr){7-8} \cmidrule(lr){9-10}
\cmidrule(lr){11-12} \cmidrule(lr){13-14} \cmidrule(lr){15-16} \cmidrule(lr){17-18}

\multicolumn{2}{c|}{\textbf{Metric}} & MSE & MAE & MSE & MAE & MSE & MAE & MSE & MAE & MSE & MAE & MSE & MAE & MSE & MAE & MSE & MAE \\

\midrule[\heavyrulewidth]

\multirow{7}{*}{\textbf{\rotatebox{90}{TimerXL}}} 

& ETTh1
& \textcolor{red}{\textbf{0.406}}	& \textcolor{red}{\textbf{0.419}}	& \textcolor{blue}{\underline{0.411}}	& 0.433
& 0.427 & 0.428
& 0.419	& \textcolor{blue}{\underline{0.427}}
& 0.438	& 0.438 & 0.422 & 0.428 & 0.458 & 0.445 & 0.454 & 0.447
\\
\cmidrule(lr){2-2} \cmidrule(lr){3-18}

& ETTh2 
& \textcolor{red}{\textbf{0.340}} & \textcolor{red}{\textbf{0.386}}	& \textcolor{blue}{\underline{0.348}}	& 0.399
& 0.349	& \textcolor{blue}{\underline{0.389}}
& 0.359	& 0.396
& 0.377 & 0.403 & 0.353 & 0.391 & 0.384 & 0.407 & 0.383 & 0.407
\\
\cmidrule(lr){2-2} \cmidrule(lr){3-18}

& ETTm1
& \textcolor{red}{\textbf{0.338}} & \textcolor{red}{\textbf{0.379}} & \textcolor{blue}{\underline{0.347}} & 0.392
& 0.377 & \textcolor{blue}{\underline{0.390}}
& 0.397	& 0.409
& 0.391 & 0.400 & 0.381 & 0.396 & 0.385 & 0.399 & 0.407 & 0.410 
\\
\cmidrule(lr){2-2} \cmidrule(lr){3-18}

& ETTm2 
& \textcolor{red}{\textbf{0.260}} & \textcolor{red}{\textbf{0.318}} & 0.268 & 0.336
& \textcolor{blue}{\underline{0.261}}	& \textcolor{blue}{\underline{0.320}}
& 0.288	& 0.347
& 0.281 & 0.326 & 0.275 & 0.322 & 0.278 & 0.325 & 0.288 & 0.332 
\\
\cmidrule(lr){2-2} \cmidrule(lr){3-18}
 
& Weather 
& \textcolor{red}{\textbf{0.225}}	& \textcolor{red}{\textbf{0.271}}	& \textcolor{blue}{\underline{0.227}}	& 0.284
& 0.228 & 0.275
& 0.255 & 0.302
& 0.252 & 0.276 & 0.243 & \textcolor{blue}{\underline{0.271}} & 0.245 & 0.276 & 0.258 & 0.278
\\

\midrule

\multirow{7}{*}{\textbf{\rotatebox{90}{Sundial}} }

& ETTh1
& \textcolor{red}{\textbf{0.395}} & \textcolor{red}{\textbf{0.419}} & 0.412 & 0.434
& \textcolor{blue}{\underline{0.411}} & 0.431
& 0.419 & 0.439 
& 0.438	& 0.438 & 0.422 & \textcolor{blue}{\underline{0.428}} & 0.458 & 0.445 & 0.454 & 0.447
\\
\cmidrule(lr){2-2} \cmidrule(lr){3-18}

& ETTh2   
& \textcolor{red}{\textbf{0.333}} & \textcolor{red}{\textbf{0.385}} & 0.347	& 0.398
& \textcolor{blue}{\underline{0.336}} & \textcolor{blue}{\underline{0.387}}
& 0.337 & 0.389  
& 0.377 & 0.403 & 0.353 & 0.391 & 0.384 & 0.407 & 0.383 & 0.407
\\
\cmidrule(lr){2-2} \cmidrule(lr){3-18}

& ETTm1 
& \textcolor{red}{\textbf{0.319}} & \textcolor{red}{\textbf{0.364}} & \textcolor{blue}{\underline{0.329}} & 0.377 
& 0.341 & \textcolor{blue}{\underline{0.374}}
& 0.340 & 0.379 
& 0.391 & 0.400 & 0.381 & 0.396 & 0.385 & 0.399 & 0.407 & 0.410
\\
\cmidrule(lr){2-2} \cmidrule(lr){3-18}

& ETTm2 
& \textcolor{red}{\textbf{0.239}} & \textcolor{red}{\textbf{0.304}} &	0.268 &	0.347	
& \textcolor{blue}{\underline{0.243}} & \textcolor{blue}{\underline{0.304}}
& 0.288 & 0.347	
& 0.281 & 0.326 & 0.275 & 0.322 & 0.278 & 0.325 & 0.288 & 0.332 
\\
\cmidrule(lr){2-2} \cmidrule(lr){3-18}

& Weather 
&  \textcolor{red}{\textbf{0.209}} & \textcolor{blue}{\underline{0.255}} & 0.212 & 0.275
& \textcolor{blue}{\underline{0.210}} & \textcolor{red}{\textbf{0.254}}
& 0.238 & 0.274
& 0.252 & 0.276 & 0.243 & 0.271 & 0.245 & 0.276 &	0.258 &	0.278
\\

\cmidrule(lr){1-18}

\multicolumn{2}{c|}{\textbf{$1^{st}$}}
& \textcolor{red}{\textbf{33}} & \textcolor{red}{\textbf{32}} & \textcolor{blue}{\underline{3}} & 0 
& \textcolor{blue}{\underline{3}} & 3
& 0 & 0 
& 0 & 0 & 1 & \textcolor{blue}{\underline{5}} & 0 & 0 &	0 & 0	
\\
\bottomrule
\end{tabular}}
\label{main results}
\end{center}
\end{table*}

\subsubsection{Forecasting Performance}\label{Forecasting Performance}

The experimental results on the public datasets are presented in Table \ref{main results}. DualWeaver consistently achieves state-of-the-art performance across all datasets, outperforming both existing encoder-decoder paradigms (AdaPTS) and native end-to-end multivariate forecasters. Our empirical evidence demonstrates that DualWeaver serves as a potent framework for unlocking the potential of Uni-TSFMs, significantly improving their multivariate forecasting performance. Specifically, DualWeaver yields notable reductions in error over vanilla Uni-TSFMs: for TimerXL, the average MSE and MAE decrease by \textbf{8.95\%} and \textbf{6.07\%}, respectively; for Sundial, the improvements reach \textbf{6.74\%} and \textbf{4.47\%}, respectively. 
Furthermore, DualWeaver outperforms full fine-tuning, reducing MSE and MAE by \textbf{3.91\%} and \textbf{1.55\%} for TimerXL, while achieving \textbf{2.67\%} and \textbf{1.12\%} for Sundial.

\subsubsection{Comparison with Multivariate TSFM} \label{Beyond Native Multivariate TSFM}
Native multivariate TSFMs, such as Chronos-2 \cite{chronos2}, are explicitly architected to capture complex cross-variable dependencies. However, this modeling capability necessitates high computation demands; specifically, by computing attention across the variable dimension, their complexity scales quadratically in the form of $O(\mathcal{V}^2)$, where $\mathcal{V}$ denotes variable count \cite{chronos2}. In contrast, Uni-TSFMs like Sundial utilize channel-independent modeling, maintaining a favorable linear complexity of $O(\mathcal{V})$.

\begin{figure}[htbp]
  \centering
  \includegraphics[width=0.98\columnwidth]{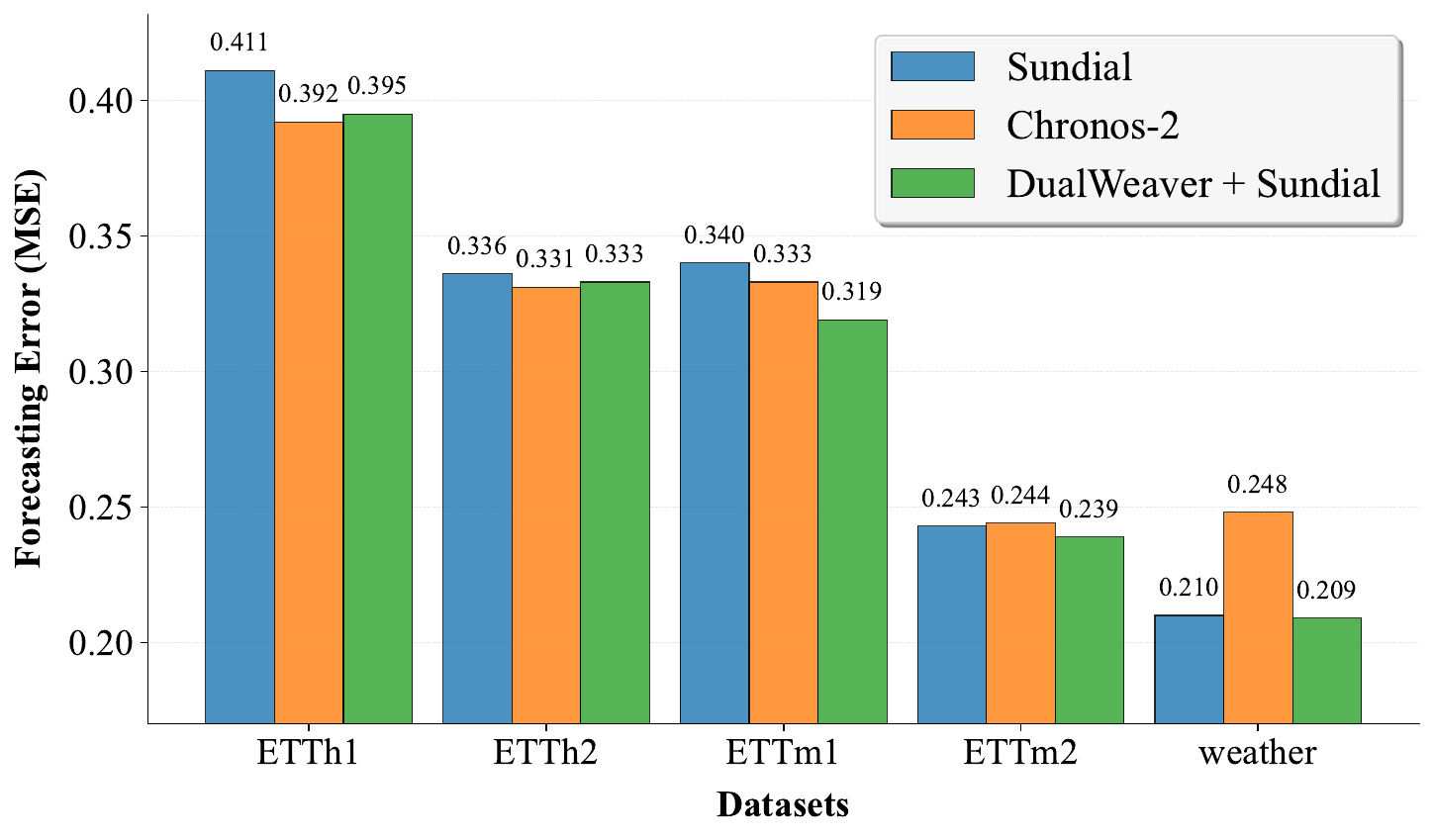}
  \caption{Performance comparison between Sundial (128M) and Chronos-2 (119M). While Chronos-2 benefits from cross-variable interactions in certain scenarios, DualWeaver enables the adapted Sundial to approach or exceed such performance.}
  \label{fig:compare}
\end{figure}

To ensure an equitable comparison, we evaluate Sundial (128M) and Chronos-2 (119M), both of which possess a comparable parameter scale. We report the optimal performance (the better of Vanilla or SFT) for each baseline to represent their strongest predictive capabilities. As illustrated in Figure \ref{fig:compare}, Chronos-2 leverages its native multivariate modeling to outperform the Sundial on most ETT subsets. Crucially, after integration with the DualWeaver framework, Sundial approaches Chronos-2's performance on ETTh and surpasses it on ETTm subsets. The results demonstrate that DualWeaver effectively bridges the gap between univariate and multivariate modeling, allowing computationally efficient Uni-TSFMs to achieve superior accuracy without the quadratic complexity of native multivariate architectures. Notably, Sundial significantly outperforms Chronos-2 on the Weather dataset; a comprehensive analysis regarding the influence of cross-variable correlation on this performance divergence is provided in Section \ref{Robustness to Irrelevant Dimensions}.

\subsubsection{Robustness to Irrelevant Dimensions} \label{Robustness to Irrelevant Dimensions}

To investigate the performance divergence in Section \ref{Beyond Native Multivariate TSFM} regarding the superior performance of the channel-independent Sundial over the native multivariate Chronos-2 on the Weather dataset, we employ the \textbf{Cross-Time Cross-Variate Correlation Coefficient} \cite{liu2024unitsteffectivelymodelinginterseries}. The metric effectively characterizes the dynamic dependencies among variables across series, with details and comprehensive results provided in Appendix \ref{Multivariate Correlations}. As shown in Table \ref{DatasetCorr}, the average correlation for the Weather dataset is remarkably low compared to the ETT series.
This sparse correlation poses a significant challenge for models that rely on deep coupling of both time and variable dimensions.
\begin{table}[htbp]
\begin{center}
\caption{Cross-Time Cross-Variate Correlation Coefficient}
\resizebox{\linewidth}{!}{
\begin{tabular}{l|c|c|c|c|c}
\toprule
{\textbf{Measurement}} &  \textbf{ETTh1} &  \textbf{ETTh2}  & \textbf{ETTm1}  & \textbf{ETTm2}  & \textbf{Weather} \\
\midrule
correlations & 0.0238 & 0.0334 & 0.0314  & 0.0420  & \textbf{0.0007} \\
\bottomrule
\end{tabular}
}
\label{DatasetCorr}
\end{center}
\end{table}

Native multivariate TSFMs and existing adapters attempt to extract predictive features through intensive modeling that deeply modifies the input $\mathbf{X}$ based on perceived cross-variable relationships.
However, in scenarios with weak cross-variable dependencies, this structural over-coupling forces such models to learn spurious relationships from irrelevant features, leading to significant generalization failure. Instead, as shown in Eq. \ref{alpha} and Eq. \ref{beta}, DualWeaver employs a residual-like surrogate mechanism that integrates multivariate insights as a supplement to the original series. Under weak-correlation conditions, this increment can adaptively decrease to avoid excessive alteration of $\mathbf{X}$. This design ensures that the model maintains a remarkably stable error profile even when cross-variable correlations are sparse, whereas deep-coupling baselines exhibit progressive performance degradation as noise increases.

\begin{figure}[htbp]
  \centering
  \includegraphics[width=0.98\columnwidth]{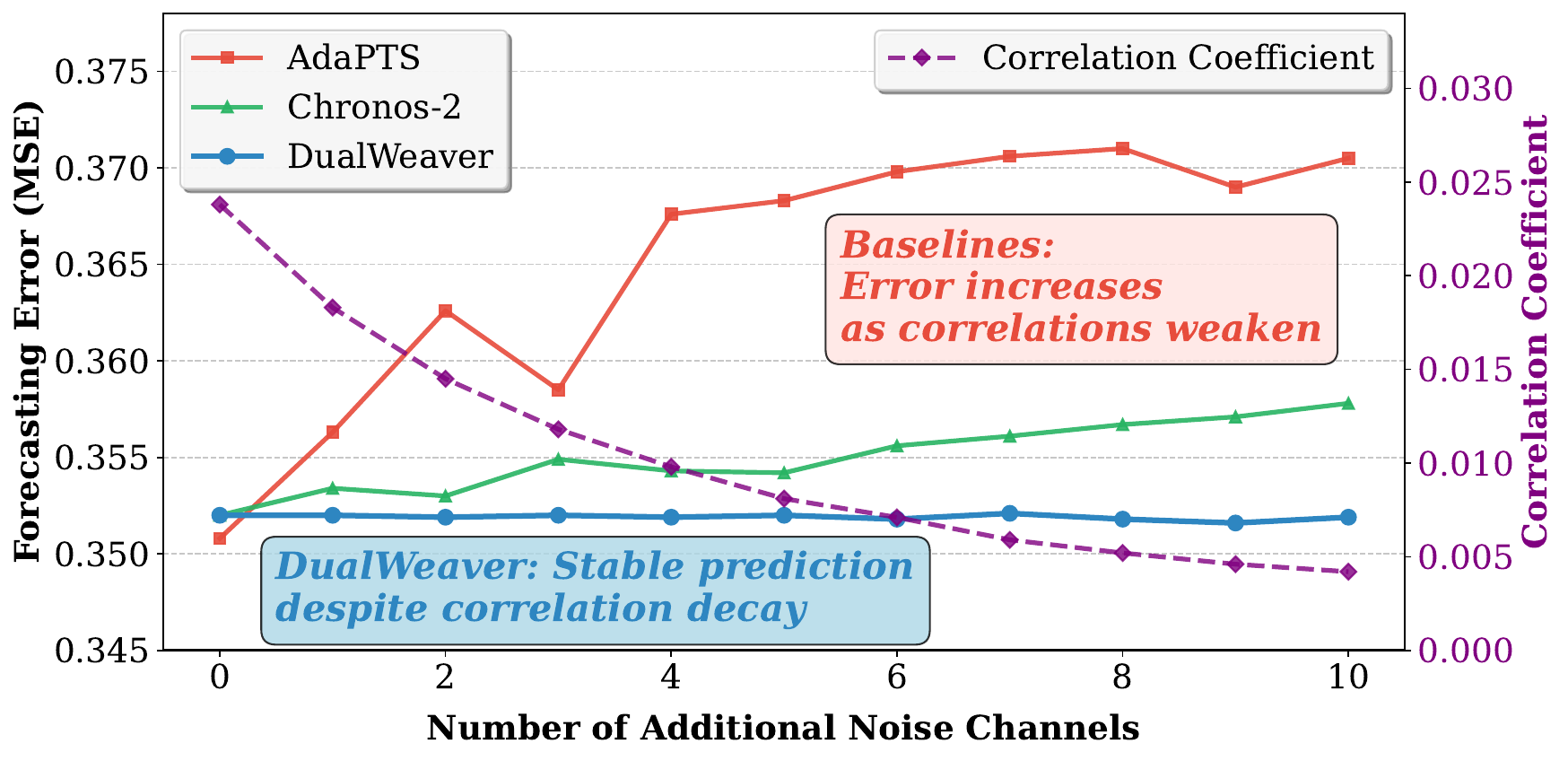}
  \caption{Predictive stability under increasing noise. DualWeaver remains robust to additional noise, outperforming baselines that degrade progressively. Done with forecasting horizon at 96.}
  \label{fig:noise}
\end{figure}

To empirically evaluate method robustness under weak cross-variable dependencies, we inject varying numbers of non-informative noise sequences sampled from a standard normal distribution $\mathcal{N}(0,1)$ into the ETTh1 dataset. As illustrated in Figure \ref{fig:noise}, while the prediction errors of Chronos-2 and AdaPTS grow progressively as the number of noise channels increases, DualWeaver maintains a remarkably stable error profile. These results underscore DualWeaver’s superior generalization capability and its effectiveness in handling real-world environments characterized by varying levels of variable correlation.

\subsubsection{Predictability in the Surrogate Space}
\begin{figure}[htbp]
  \centering
  \includegraphics[width=0.93\columnwidth]{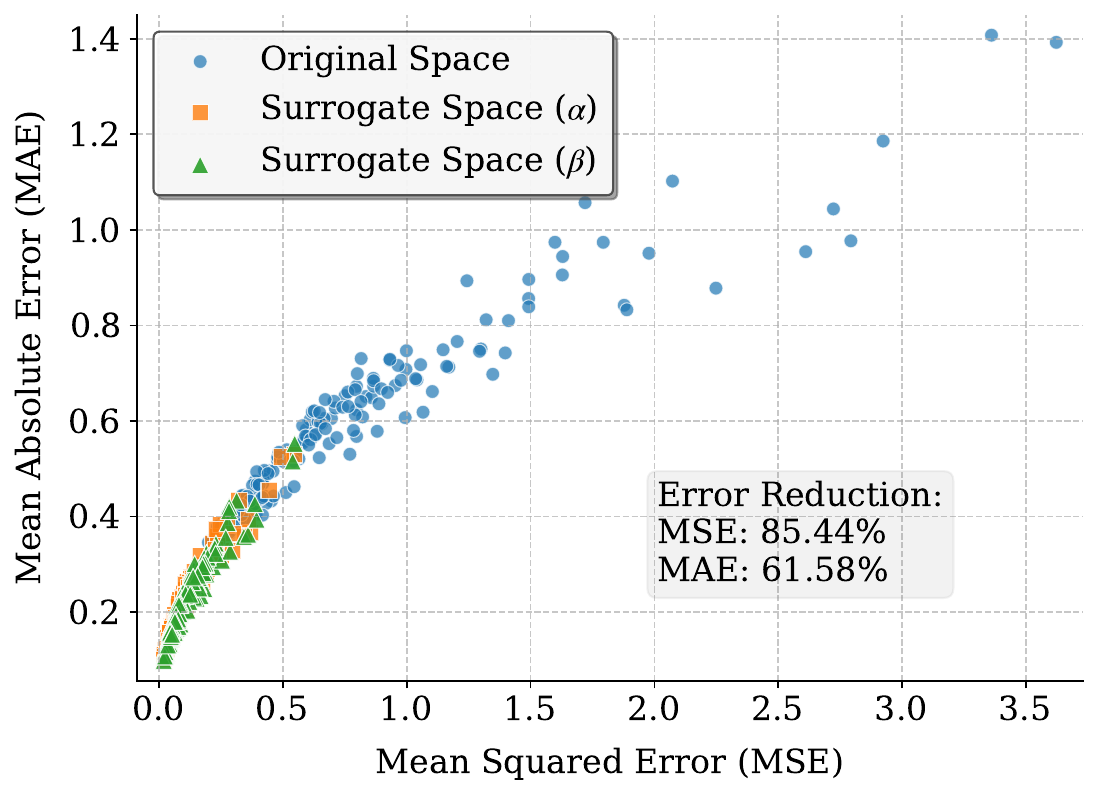}
  \caption{Predictability enhancement in the surrogate space. DualWeaver transforms complex multivariate signals into predictable surrogates, yielding an average reduction of \textbf{85.44\%} in MSE and \textbf{61.58\%} in MAE compared to the original input space.}
  \label{fig:error_distribution}
\end{figure}

DualWeaver enhances Uni-TSFMs for multivariate forecasting by projecting multivariate inputs into a highly predictable surrogate space. 
To quantify this enhancement, we compare the forecasting errors of the original ETTh1 time series with those of our surrogates, both using the same frozen TimerXL foundation model. Figure \ref{fig:error_distribution} illustrates the resulting error distributions after adaptation convergence, with each data point representing the local error mean computed over a 100-timestep rolling window to enhance clarity.

Empirically, the surrogates exhibit significantly lower error profiles, achieving average reductions of \textbf{85.44\%} in MSE and \textbf{61.58\%} in MAE. These substantial improvements provide concrete evidence that our surrogate mechanism effectively simplifies complex multivariate signals into more tractable representations, thereby effectively aligning the robust temporal modeling of Uni-TSFMs with the requirements of multivariate forecasting.

\subsection{Computational Efficiency} \label{Computational Efficiency}

\begin{figure}[htbp]
  \centering
  \includegraphics[width=0.95\columnwidth]{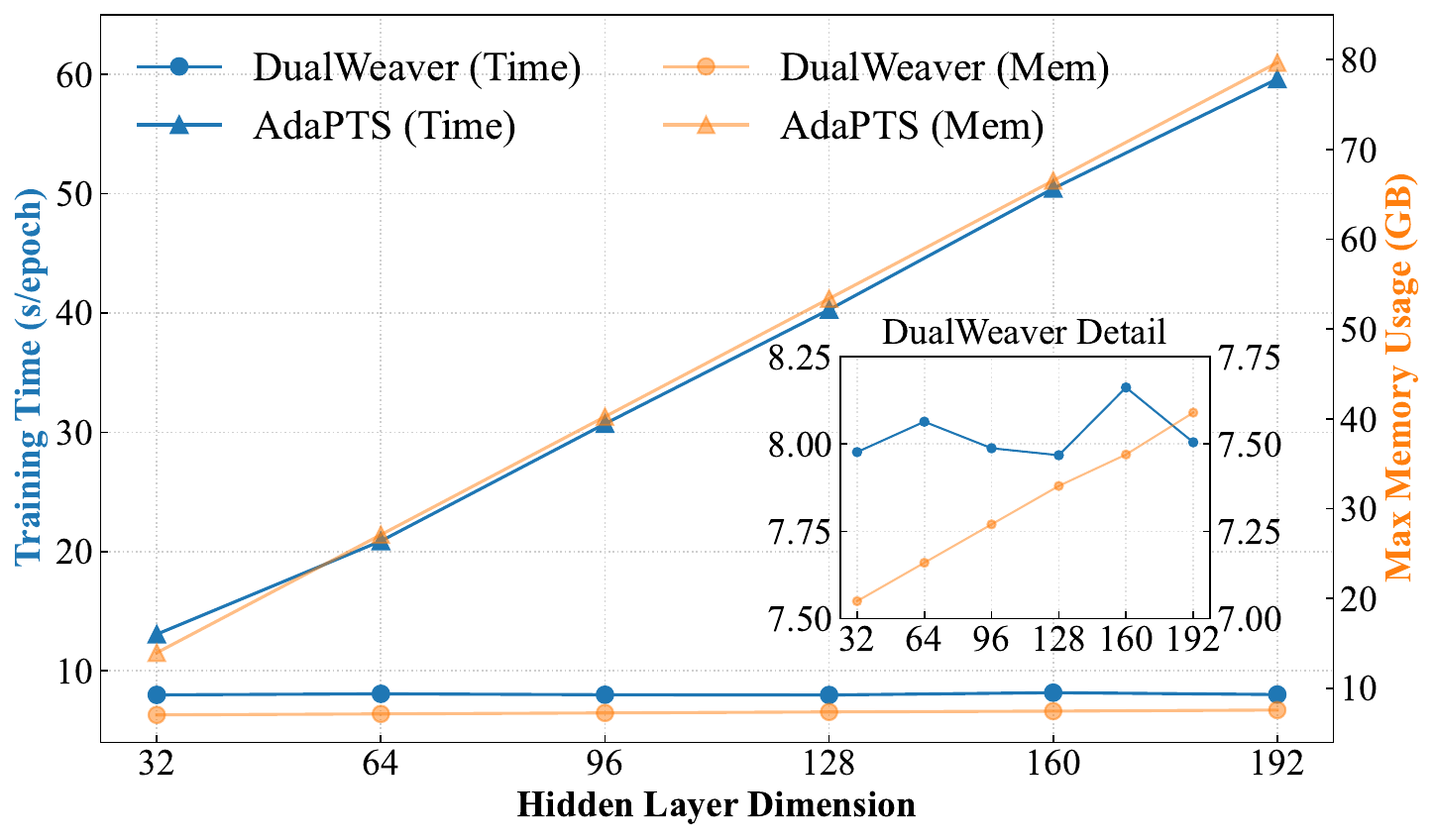}
  \caption{Training efficiency and scalability comparison by training time per epoch (s) and peak memory consumption (GB) on the ETTh1 dataset using TimerXL. While AdaPTS exhibits linear growth in computational costs, DualWeaver maintains a near-constant resource footprint across varying hidden dimensions.}
  \label{fig:efficiency}
\end{figure}

This section compares the training efficiency and memory consumption of DualWeaver with those of AdaPTS.
Benchmarks were performed on the ETTh1 dataset across varying hidden dimensions. As illustrated in Figure~\ref{fig:efficiency}, our proposed DualWeaver maintains low training time cost and memory footprint across varying hidden dimensions. At the same time, AdaPTS exhibits near-linear growth in both metrics as the number of hidden dimensions increases. This scalability limitation stems from AdaPTS's architectural design: it projects input features to match the hidden dimension size before processing them through the foundation model, causing computational costs to scale directly with hidden dimension size.
Consequently, practitioners must trade off efficiency against modeling capacity, significantly limiting practical deployment.
In contrast, DualWeaver preserves dimensionality alignment between input features and the foundation model's expected inputs. This design principle ensures that computational efficiency remains nearly constant as the number of hidden dimensions increases. 

\subsection{Ablation Study}
\label{Ablation_study}

\subsubsection{Robustness from Error Bound Condition}\label{Robustness from Error Bound Condition}
\begin{figure}[!htb]
  \centering
  \includegraphics[width=0.98\columnwidth]{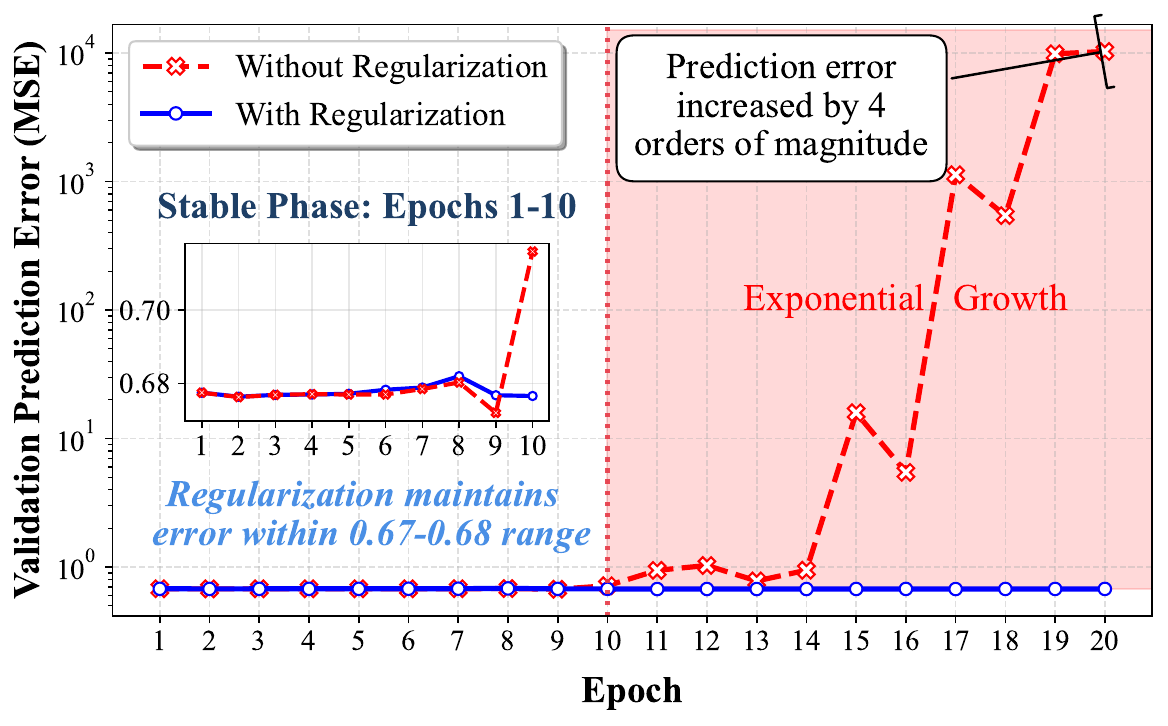}
  \caption{Impact of the regularization on training stability. Without the theoretical constraint (Eq.~\ref{condition}), the model experiences catastrophic divergence after epoch 10, with validation error increasing by four orders of magnitude. Using our proposed regularization ensures consistent stability and performance.}
  \label{fig:re}
\end{figure}

To validate the theoretical error bound guarantee provided by Eq.~\ref{condition}, we examine the impact of the proposed regularization on convergence stability. Figure \ref{fig:re} contrasts the validation error trajectories on ETTh1 using TimerXL with and without this constraint, illustrating its role in ensuring robust adaptation. The empirical results reveal a stark divergence in training stability. Under error-bound regularization, DualWeaver exhibits a remarkably stable validation MSE. In contrast, the unregularized variant, while initially matching the stable performance during the first 10 epochs, undergoes a catastrophic collapse thereafter. Specifically, the validation error enters an exponential growth regime, increasing by over 4 orders of magnitude. 
The proposed regularization effectively serves as a safety boundary, preserving the integrity of the surrogate space and ensuring robust adaptation even during extended exploration.

\subsubsection{Dimensional Scalability}\label{Dimensional Scalability}
To further evaluate the robustness of DualWeaver when handling higher-dimensional data, we conducted scalability experiments on the ECL dataset. Featuring 321 variables, this dataset provides a rigorous benchmark for assessing model performance in large-scale multivariate scenarios. By selecting the first $N$ variables ($N \in \{5, 10, 20, ..., 321\}$), we constructed subsets of varying dimensions to observe the performance gains of DualWeaver and AdaPTS when adapting the TimerXL foundation model.

\begin{figure}[htbp]
  \centering
  \includegraphics[width=\columnwidth]{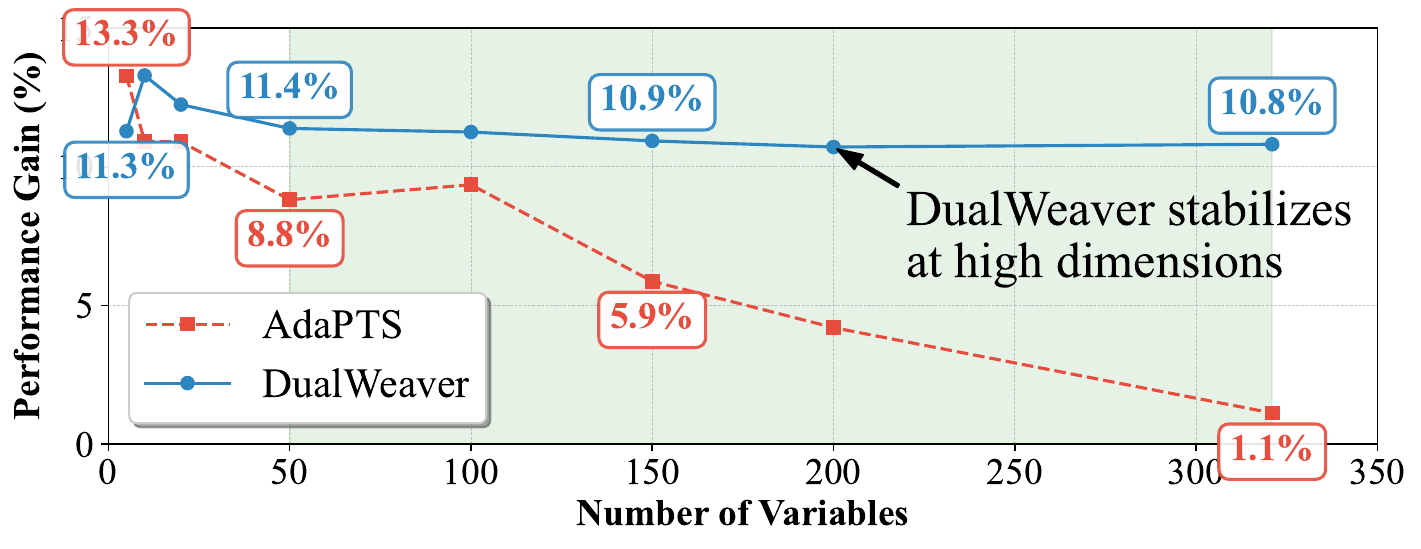}
  \caption{Dimensionality Scalability on ECL Dataset. Comparison of performance gains (MSE reduction) achieved by DualWeaver and AdaPTS on TimerXL as the number of variables $N$ increases. While AdaPTS exhibits sensitivity to increasing dimensionality with declining returns, DualWeaver demonstrates superior robustness and stable predictive gains even as $N$ reaches 321.}
  \label{fig:ecl}
\end{figure}

The experimental results are illustrated in Figure \ref{fig:ecl}. At low variable counts ($N \le 20$), both adaptation paradigms yield comparable and significant improvements in accuracy over the foundation model. However, as $N$ increases, the marginal utility of AdaPTS progressively decreases. This performance degradation stems from its underlying encoder-decoder architecture; as dimensionality scales, a substantial portion of the adaptation capacity is consumed by reconstructing the original variables from latent representations, rather than by distilling critical cross-variable dependencies. In contrast, DualWeaver maintains a remarkably consistent performance gain across the entire range of variable scales. This resilience is attributed to our dual-surrogate mechanism, which leverages a shared feature-fusion module to optimize directly within the surrogate space. Eliminating parametric reconstruction enables DualWeaver to focus solely on capturing cross-variable dependencies, thereby contributing to its exceptional scalability and stable performance in high-dimensional settings.

\section{Conclusion}
In this paper, we introduce DualWeaver, a novel framework that enables multivariate forecasting with Uni-TSFMs. It employs an innovative dual-surrogate paradigm: a learnable feature-fusion module extracts cross-variable dependencies to generate a pair of highly predictable surrogate series, enabling non-parametric reconstruction of final predictions. Unlike encoder-decoder approaches, DualWeaver eliminates capacity loss through costless, direct surrogate mapping. Supported by theoretical error-bound analysis and a novel regularization term for stable adaptation, the framework demonstrates state-of-the-art accuracy, superior computational efficiency, and robustness against noise and high-dimensional data in extensive experiments.


\section*{Impact Statement}

This paper presents a novel framework to address the efficiency and robustness challenges of TSFM in multivariate forecasting via an innovative dual-surrogate paradigm. 
By achieving state-of-the-art accuracy with superior computational efficiency and inherent robustness to noise, our method provides both a practical tool for real-world applications and a valuable conceptual advance—the learnable feature-fusion module and costless surrogate mapping—for the broader deep learning research community. 
This work is strictly focused on the scientific and engineering problem of forecasting methodology. We foresee no direct ethical risks arising from the framework itself and are committed to its responsible development and application.

\bibliography{example_paper}
\bibliographystyle{icml2026}

\newpage

\appendix

\section{Implementation Details} \label{Implementation Details}
\subsection{Datasets} \label{appendix:Datasets}

\subsubsection{Dataset Descriptions} \label{appendix:Public Dataset Descriptions}

We conduct experiments on six public datasets to evaluate the performance of the proposed DualWeaver, including: (1) ETT \cite{zhou2021informer} contains seven factors of electricity transformer from July 2016 to July 2018. There are four subsets where ETTh1 and ETTh2 are recorded every hour, and ETTm1 and ETTm2 are recorded every 15 minutes. (2) Weather\footnote{https://www.bgc-jena.mpg.de/wetter/} includes 21 meteorological factors collected every 10 minutes from the Weather Station of the Max Planck Biogeochemistry Institute in 2020. (3) Electricity \cite{wu2021autoformer} records the hourly electricity consumption data of 321 clients.

To ensure fair benchmarking against baseline methods, we strictly adhere to the identical training-validation-test split strategy used across all baseline datasets. The details of the datasets are provided in Table \ref{Dataset details}. Dim denotes the variable number of each dataset. Dataset Size denotes the total number of time points in the (train, validation, test) split, respectively. Frequency represents the sampling interval of time points.
\begin{table}[htbp]
\begin{center}
\caption{Detailed dataset descriptions.}
\resizebox{\linewidth}{!}{
\begin{tabular}{l|c|c|c}
\toprule
{\textbf{Dataset}} &  \textbf{Dim} &  \textbf{Dataset Size} & \textbf{Frequency} \\
\midrule[\heavyrulewidth]
ETTh1 & 7 & (8545, 2881, 2881) & Hourly \\
\midrule
ETTh2 & 7 & (8545, 2881, 2881) & Hourly \\
\midrule
ETTm1 & 7 & (34465, 11521, 11521) & 15min \\
\midrule
ETTm2 & 7 & (34465, 11521, 11521) & 15min \\
\midrule
Weather & 21 & (36792, 5271, 10540) & 10min \\
\midrule 
ECL & 321 & (18317, 2633, 5261) & Hourly \\
\bottomrule
\end{tabular}
}
\label{Dataset details}
\end{center}
\end{table}

\subsubsection{Data Preprocessing}
To ensure experimental consistency and align with established research benchmarks, we standardized all experimental datasets during preprocessing. Specifically, we applied StandardScaler to normalize feature scales to ensure consistency across all baseline methods and the proposed DualWeaver method.

\subsection{Training Configurations}  
\label{appendix:Training Configurations}

All experiments were implemented in PyTorch and executed on 8 NVIDIA H20 (96GB) GPUs. We used distributed data parallelism via \texttt{torchrun} to accelerate both training and inference. We adopted the AdamW optimizer~\cite{adamw} with hyperparameters $\beta_1 = 0.9$, $\beta_2 = 0.95$, and weight decay $= 1\times10^{-3}$, coupled with a cosine annealing learning rate scheduler~\cite{scheduler} configured with $T_{\text{max}} = 10$ and $\eta_{\text{min}} = 1\times10^{-8}$. All models underwent 10 training epochs with early stopping (patience $= 3$). For adapter-based methods (including DualWeaver and AdaPTS), we froze all Uni-TSFM parameters. Multivariate forecasting experiments used a fixed batch size of 32, with gradient accumulation and micro-batching to maintain equivalent batch processing while preventing out-of-memory errors. In contrast, univariate fine-tuning utilized a fixed batch size of 256 with sequential variable loading. Across all configurations, we preserved complete test set integrity by omitting the \enquote{drop last} strategy.

We adopted a context length of 2880 for Uni-TSFMs, consistent with the configurations used in Sundial \cite{liu2025sundial}. This extended context length is crucial for fully activating the inherent long-term dependency modeling capabilities of foundation models. 
For the multivariate forecasters (TimePro, SimpleTM, TimeMixer, and iTransformer), we directly report their published results as presented in their original papers, maintaining the input length of 96 as specified in their established experimental protocols. 

For all experiments, we use a grid search to optimize hyperparameters. For all adapter-based methods (DualWeaver and AdaPTS), we search over learning rates $\{10^{-1}, 10^{-2}, 10^{-3}, 10^{-4}\}$. For full fine-tuning, we utilize reduced learning rates $\{5\times10^{-5}, 10^{-5}, 5\times10^{-6}, 10^{-6}\}$ to mitigate the risk of overfitting. For AdaPTS, we adopt its representative variant, LinearAE, as the baseline. Here, the hidden dimension ($d_{\text{hidden}}$) is defined by the output channels of the first linear projection layer in both AdaPTS and DualWeaver, which characterizes the models' capability to embed multivariate features into a latent space. The hidden layer dimension for both AdaPTS and our proposed DualWeaver is configured according to the widely adopted benchmark setting established by TimesNet \cite{wu2022timesnet}. This dimension is dynamically adjusted based on the number of variables ($\mathcal{V}$) in each dataset, balancing model capacity and computational efficiency. The calculation is defined by:

\begin{equation}
d_{\text{hidden}} = \min \left\{ \max \left\{ 2^{\lceil \log_2 \mathcal{V} \rceil}, 32 \right\}, 512 \right\}
\end{equation}

This design ensures a baseline model complexity that scales with the intrinsic dimensionality of the data while preventing the hidden layer from becoming impractically small or excessively large.

\subsection{Adaptation Pipeline}
To ensure a fair comparison, our adaptation process strictly follows the stage-wise adaptation paradigm established by the baseline AdaPTS \cite{benechehab_adapts_2025}. Specifically, the Uni-TSFM is first optimized via methods such as full fine-tuning, linear probing \cite{goswami2024momentfamilyopentimeseries}, or LoRA \cite{hu2021loralowrankadaptationlarge}; in this work, we adopt full fine-tuning as the primary strategy. Following this initial stage, all parameters of the Uni-TSFM backbone are frozen, while only those within the DualWeaver framework are trained.

\subsection{Design and Implementation}

\subsubsection{Geometric Interpretation of Dual Surrogates} \label{geometric}

The dual surrogates can be understood as constructing a pair of learnable tangent vectors on the representation manifold \(\mathcal{M}_{\text{repr}}\). At the base point \(f(\mathbf{X})\), we define two adaptive exploration directions:
\[
\mathbf{v}_\alpha = \mathbf{w}_\alpha \odot \mathbf{X}, \quad \mathbf{v}_\beta = \mathbf{w}_\beta \odot \mathbf{X},
\]
where the channel-wise weights \(\mathbf{w}_\alpha, \mathbf{w}_\beta \in \mathbb{R}^C\) are initialized to 1 and optimized during adaptation.

\textbf{Initialization}: Setting \(\mathbf{w}_\alpha = \mathbf{w}_\beta = \mathbf{1}\) establishes a symmetric, unbiased starting point, assuming a preliminary alignment between the raw input space and the tangent space of the representation manifold.

\textbf{Optimization}: The learning process calibrates these tangent vectors to the specific local geometry of the target domain. Crucially, \(\mathbf{w}_\alpha\) and \(\mathbf{w}_\beta\) are learned independently. This design allows the model to discover that the most informative directions for feature enhancement (\(\mathbf{v}_\alpha\)) and suppression (\(\mathbf{v}_\beta\)) are not necessarily symmetric around \(f(\mathbf{X})\). It captures the anisotropic nature of the local data distribution---the manifold may curve more sharply or contain more signal in certain directions.

\textbf{Theoretical Advantage}: Thus, the dual surrogates \(\mathbf{S}_\alpha = f(\mathbf{X}) + \mathbf{v}_\alpha\) and \(\mathbf{S}_\beta = f(\mathbf{X}) - \mathbf{v}_\beta\) do not merely perform fixed affine transformations. Instead, they implement a data-driven, structured exploration of the neighborhood of the representation manifold, guided by the adapted tangent vectors \(\mathbf{v}_\alpha\) and \(\mathbf{v}_\beta\). This provides a principled and efficient mechanism for domain adaptation that is both flexible (due to learning) and regularized (by the tangent space constraint).

\subsubsection{Uni-TSFMs}
For the Uni-TSFMs Sundial and TimerXL, we use their original model architectures and the officially released pre-trained weights from the Huggingface platform\footnote{https://huggingface.co} to ensure experimental reproducibility.
\subsubsection{Multivariate Feature-Fusion Module}
\label{appendix:Multivariate Feature-Fusion Module}
The DualWeaver framework incorporates a learnable feature-fusion module $f(\cdot): \mathbb{R}^{T \times C} \rightarrow \mathbb{R}^{T \times C}$ to capture cross-variable dependencies in multivariate time series data.
This module is implemented as a multi-layer perceptron (MLP), which consists of two fully-connected layers: the first layer projects the input features to a hidden dimension, followed by a SiLU activation and dropout (with a rate of 0.1) for nonlinear transformation; the second layer then projects the features back to the original dimensionality. The weights of the final linear layer are initialized to zero, ensuring that the optimization process begins from the Uni-TSFM's original predictions.

\section{DualWeaver on Vanilla Uni-TSFMs}
\label{appendix:DualWeaver on vanilla Uni-TSFMs}

The non-intrusive architecture of DualWeaver enables seamless compatibility with diverse Uni-TSFM implementations. Leveraging this flexibility, we conduct an extended performance assessment of DualWeaver applied to vanilla Uni-TSFMs. 
As illustrated in Table \ref{vanilla_results}, DualWeaver consistently outperforms AdaPTS across the vast majority of evaluation metrics when integrated with vanilla Uni-TSFMs. Notably, on the Sundial model, DualWeaver achieves a substantial performance margin over the AdaPTS baseline. Specifically, compared to zero-shot forecasting, our framework yields a 3.70\% reduction in MSE and a 2.26\% reduction in MAE. These results underscore the superior generalizability of the dual-surrogate mechanism, demonstrating its ability to effectively unlock the multivariate potential of frozen foundation models without the parametric reconstruction overhead typical of prior encoder-decoder architectures.

\begin{table}[htbp]
\begin{center}
\caption{Multivariate forecasting results on \underline{\textbf{vanilla Uni-TSFMs}} with forecasting horizons $\in \{96, 192, 336, 720\}$. Averaged results are reported here with the best in \textbf{bold}.}
\label{vanilla_results}

\resizebox{\linewidth}{!}{
\begin{tabular}{c|c|cc|cc|cc}
\toprule

\multicolumn{2}{c|}{\textbf{Methods}} 
& \multicolumn{2}{c}{\textbf{DualWeaver}} 
& \multicolumn{2}{c}{\textbf{AdaPTS}} 
& \multicolumn{2}{c}{\textbf{Vanilla}} \\

\cmidrule(lr){1-8}

\multicolumn{2}{c|}{\textbf{Metric}} & MSE & MAE & MSE & MAE & MSE & MAE \\

\cmidrule(lr){1-8}

\multirow{6}{*}{\textbf{\rotatebox{90}{TimerXL}} }

& ETTh1 
& 0.409	& \textbf{0.418} & \textbf{0.407} & 0.429	& 0.419	& 0.427 \\
\cmidrule(lr){2-2} \cmidrule(lr){3-8} 

& ETTh2 
& \textbf{0.341}	& \textbf{0.386}	& 0.344	& 0.397	& 0.359	& 0.396 \\
\cmidrule(lr){2-2} \cmidrule(lr){3-8} 

& ETTm1
& 0.368	& \textbf{0.397}	& \textbf{0.360} & 0.400 & 0.397	& 0.409 \\
\cmidrule(lr){2-2} \cmidrule(lr){3-8} 

& ETTm2 
& 0.276	& \textbf{0.335}	& \textbf{0.273}	& 0.344 & 0.288	& 0.347 \\
\cmidrule(lr){2-2} \cmidrule(lr){3-8} 

& Weather 
& 0.245	& \textbf{0.289}	& \textbf{0.244}	& 0.304 & 0.255	& 0.302 \\

\midrule

\multirow{7}{*}{\textbf{\rotatebox{90}{Sundial}} }

& ETTh1 
& \textbf{0.403}	& \textbf{0.426}	& 0.412	& 0.435	& 0.419	 &0.439 \\
\cmidrule(lr){2-2} \cmidrule(lr){3-8} 

& ETTh2
& \textbf{0.332}	& \textbf{0.385}	& 0.347	& 0.399	& 0.337	& 0.389\\
\cmidrule(lr){2-2} \cmidrule(lr){3-8} 

& ETTm1
& \textbf{0.329}	& \textbf{0.373}	& 0.349	& 0.397	& 0.340	& 0.379 \\
\cmidrule(lr){2-2} \cmidrule(lr){3-8} 

& ETTm2
& \textbf{0.256}	& \textbf{0.321}	 & 0.268	& 0.344	& 0.261	& 0.323\\
\cmidrule(lr){2-2} \cmidrule(lr){3-8} 

& Weather
& 0.230	& \textbf{0.271}	& \textbf{0.226}	& 0.289	& 0.238	& 0.274 \\

\bottomrule
\end{tabular}

}
\end{center}
\end{table}

\section{Extensibility}
\label{Extensibility}

As shown in benchmarks like BasicTS+ \cite{shao_exploring_2025}, the strength of cross-variable correlations varies significantly across datasets. Models such as DLinear \cite{zeng2023transformers}, which rely on simple linear architectures, tend to perform well on datasets with weak dependencies, while more complex GNN-based approaches excel when strong cross-variable relationships are present. To accommodate such diversity in real-world data, the feature-fusion module in DualWeaver is designed to be highly extensible. As long as the output shape matches the input multivariate sequence, the feature-fusion module can be replaced with any neural network architecture—such as CNNs, GNNs, or Transformers—enabling the model to adapt to varying correlation strengths without altering the overall framework.

In this section, we define a CNN-based feature-fusion module that employs 1D convolutional layers applied along all the dimensions, treating the variables as channels of convolution:
\begin{equation}
f(\mathbf{X})_{\text{CNN}}^{j,t} = \sum_{i=1}^{C} \left( \mathbf{X}_i \circledast \mathbf{W}_{j,i} \right)_t + \mathbf{b}_j
\end{equation}
where $i$ is the index of the input channel, $j$ is the index of the output channel, $\circledast$ is the convolution operation, $\mathbf{X}_i\in \mathbb{R}^{L}$ is the time series of the $i$-th variable, $\mathbf{W}_{j,i} \in \mathbb{R}^{K}$ is the kernel weight matrices for the $i$-th input channel and the $j$th output channel, and $\mathbf{b}_j \in \mathbb{R}$ is the $j$th bias for the output. Here, for simplicity, we present the CNN-based module using only one convolutional layer. In experiments, the CNN-based variant employs two 1D convolutional blocks (kernel size 5, stride 1, replicate padding) where the initial convolution expands the input dimensionality to a configurable hidden dimension, followed by layer normalization, SiLU activation, and dropout (p=0.1). The CNN-based feature-fusion module uses convolutional layers to capture cross-variable dependencies within a specified time window automatically, leveraging parameter sharing and locality to reduce computational cost. 


We evaluate the multivariate forecasting performance of the CNN-based DualWeaver across five benchmark datasets. Following the experimental protocols detailed in Appendix \ref{Implementation Details}, our empirical results (Table \ref{WeaverCNN}) demonstrate that this variant consistently achieves substantial precision gains across all evaluated scenarios. Specifically, DualWeaver yields average reductions in MSE and MAE of \textbf{8.70\%} and \textbf{5.99\%} for TimerXL, and \textbf{6.68\%} and \textbf{4.47\%} for Sundial, respectively. Furthermore, the inherent extensibility of the feature-fusion module underscores the framework's versatility and broad applicability across diverse real-world tasks.

\begin{table}[htbp]
\begin{center}
\caption{Multivariate forecasting results of CNN-variant DualWeaver with forecasting horizons $\in \{96, 192, 336, 720\}$. Averaged results are reported here with the best in \textbf{bold}.}
\label{WeaverCNN}

\resizebox{\linewidth}{!}{
\begin{tabular}{c|c|cc|cc|cc}
\toprule

\multicolumn{2}{c|}{\textbf{Methods}} 
& \multicolumn{2}{c}{\textbf{WeaverCNN}} 
& \multicolumn{2}{c}{\textbf{AdaPTS}} 
& \multicolumn{2}{c}{\textbf{Vanilla}} \\

\cmidrule(lr){1-2} \cmidrule(lr){3-4} \cmidrule(lr){5-6}
\cmidrule(lr){7-8}  

\multicolumn{2}{c|}{\textbf{Metric}} & MSE & MAE & MSE & MAE & MSE & MAE \\

\cmidrule(lr){1-8}

\multirow{6}{*}{\textbf{\rotatebox{90}{TimerXL}} }

& ETTh1 
& \textbf{0.406}	& \textbf{0.419} &	0.411	& 0.433	& 0.419	& 0.427 \\
\cmidrule(lr){2-2} \cmidrule(lr){3-8} 

& ETTh2 
& \textbf{0.342}	& \textbf{0.386}	& 0.348	& 0.39	& 0.359	& 0.396 \\
\cmidrule(lr){2-2} \cmidrule(lr){3-8} 

& ETTm1
& \textbf{0.336}	& \textbf{0.388}	& 0.347	& 0.392	& 0.397	& 0.409 \\
\cmidrule(lr){2-2} \cmidrule(lr){3-8} 

& ETTm2 
& \textbf{0.260}	& \textbf{0.318}	& 0.268	& 0.336	& 0.288	& 0.347 \\
\cmidrule(lr){2-2} \cmidrule(lr){3-8} 

& Weather 
& \textbf{0.228}	& \textbf{0.273}	& 0.227	& 0.284	& 0.255	& 0.302 \\

\midrule

\multirow{7}{*}{\textbf{\rotatebox{90}{Sundial}} }

& ETTh1 
& \textbf{0.395}	& \textbf{0.419}	& 0.412	& 0.434	& 0.419	 &0.439 \\
\cmidrule(lr){2-2} \cmidrule(lr){3-8} 

& ETTh2
& \textbf{0.333}	& \textbf{0.385}	& 0.347	& 0.398	& 0.337	& 0.389\\
\cmidrule(lr){2-2} \cmidrule(lr){3-8} 

& ETTm1
& \textbf{0.320}	& \textbf{0.364}	& 0.329	& 0.333	& 0.340	& 0.379 \\
\cmidrule(lr){2-2} \cmidrule(lr){3-8} 

& ETTm2
& \textbf{0.239}	& \textbf{0.304}	 & 0.268	& 0.307	& 0.261	& 0.323\\
\cmidrule(lr){2-2} \cmidrule(lr){3-8} 

& Weather
& \textbf{0.209}	& \textbf{0.255}	& 0.212	& 0.275	& 0.238	& 0.274 \\

\bottomrule
\end{tabular}

}
\end{center}
\end{table}

\section{Evaluation on Additional Datasets}
\label{Evaluation on More Datasets}

To further validate the generalizability of DualWeaver in enhancing Uni-TSFMs for multivariate forecasting, we evaluate its performance using TimerXL across several additional datasets. This expanded evaluation encompasses the Exchange \cite{wu2021autoformer}, Solar-Energy \cite{lai2018modelinglongshorttermtemporal}, and IEAC (Industrial Equipment Air Conditioning Dataset)—a real-world industrial dataset collected from a production environment. Exchange collects the panel data of daily exchange rates from 8 countries from 1990 to
2016. Solar-Energy records solar power production from 137 PV plants in 2006, sampled at 10-minute intervals. IEAC is from an HVAC (Heating, Ventilation, and Air Conditioning) system, which comprises five critical parameters: (1) HVAC1 (System Operating Mode): A discrete variable indicating the current operational state/phases. (2) HVAC2 (Compression Power): A continuous variable representing the power level of the upstream air sources, set by system demand (higher values = greater air source pressure). (3) HVAC3 (Distribution Duct Pressure): A continuous variable directly reflecting the pressure within the distribution lines, primarily driven by HVAC2. (4) HVAC4 (User Pressure): A continuous variable denoting the pressure at downstream user systems, influenced by HVAC3 and internal system regulation. (5) HVAC5 (User Flow Rate): A continuous variable indicating the flow rate to downstream user systems, affected by HVAC3 and internal system regulation. Key relationships show a causal chain: HVAC2 determines HVAC3, which subsequently influences both downstream HVAC4 and HVAC5, subject to the HVAC system's active control. HVAC1 governs system state while performance is regulated via continuous monitoring of pressures (HVAC3 and HVAC4) and flow (HVAC5). The detailed dataset splits are summarized in Table \ref{Dataset additional details}. Given the limited scale of the Exchange dataset, we employ a reduced context length of 512 time points and shorter forecast horizons of $\{24, 36, 48, 60\}$ steps. This configuration ensures sufficient training and testing samples while maintaining statistical reliability. Several additional multivariate public datasets, such as Traffic (PEMS\footnote{http://pems.dot.ca.gov/}), were excluded from evaluation because they are already included in the pre-training datasets used for these Uni-TSFMs.

\begin{table}[htbp]
\begin{center}
\caption{Detailed additional dataset descriptions.}
\resizebox{\linewidth}{!}{
\begin{tabular}{l|c|c|c}
\toprule
{\textbf{Dataset}} &  \textbf{Dim} &  \textbf{Dataset Size} & \textbf{Frequency} \\
\midrule[\heavyrulewidth]
Exchange & 8 & (5120, 665, 1422) & Daily \\
\midrule
Solar-Energy & 137 & (36601, 5161, 10417) & 10 min\\
\midrule
IEAC & 5 & (70000, 10000, 20000) & 5s \\
\bottomrule
\end{tabular}
}
\label{Dataset additional details}
\end{center}
\end{table}

Performance results for the Exchange, Solar-Energy, and IEAC datasets are summarized in Table \ref{additional_datasets}. DualWeaver achieves state-of-the-art (SOTA) performance in five out of the six evaluated metrics, with AdaPTS marginally outperforming it only on the MAE metric for the Solar-Energy dataset. The robust forecasting performance across these additional real-world datasets underscores DualWeaver's superior generalization capabilities and its potential for deployment in diverse industrial environments.

\begin{table}[htbp]
\begin{center}
\caption{Multivariate forecasting on additional datasets with forecasting horizons $\in \{24, 36, 48, 60\}$ for Exchange and $\in \{96, 192, 336, 720\}$ for others. Averaged results are reported here with the best in \textbf{bold}.}
\label{additional_datasets}

\resizebox{\linewidth}{!}{
\begin{tabular}{c|cc|cc|cc|cc}
\toprule

\textbf{Methods}
& \multicolumn{2}{c}{\textbf{DualWeaver}} 
& \multicolumn{2}{c}{\textbf{AdaPTS}} 
& \multicolumn{2}{c}{\textbf{SFT}} 
& \multicolumn{2}{c}{\textbf{Vanilla}} \\

\cmidrule(lr){1-1} \cmidrule(lr){2-3} \cmidrule(lr){4-5} \cmidrule(lr){6-7} \cmidrule(lr){8-9}

\textbf{Metric} & MSE & MAE & MSE & MAE & MSE & MAE & MSE & MAE \\

\cmidrule(lr){1-9}

Exchange
& \textbf{0.055}	& \textbf{0.158} &	0.057	& 0.163	& 0.057	& 0.160	& 0.064	& 0.168 \\
\cmidrule(lr){1-1} \cmidrule(lr){2-9} 

Solar-Energy
& \textbf{0.354} &	0.365	& 0.376	& \textbf{0.359} &	0.403	& 0.361	& 0.517	& 0.492 \\
\cmidrule(lr){1-1} \cmidrule(lr){2-9} 

IEAC
& \textbf{0.750} & \textbf{0.439} & 0.754 & 0.441 & 0.767 & 0.479 & 0.930 & 0.562 \\

\bottomrule
\end{tabular}
}
\begin{center}
\end{center}
\end{center}
\end{table}


\section{Multivariate Correlations}\label{Multivariate Correlations}

The \textbf{Cross-Time Cross-Variate Correlation Coefficient} \cite{liu2024unitsteffectivelymodelinginterseries} quantifies dynamic relationships between variates across asynchronous periods. By computing covariance between standardized subsequences from different variates in offset time windows, this metric captures delayed dependencies that traditional synchronous correlation coefficients cannot reveal, providing a precise characterization of asynchronous cross-variate interactions essential for modeling complex multivariate correlations. According to it, we measure the multivariate correlations as follows:
\begin{equation}
    R^{(i,j)}(t,t',P) 
    = \frac{1}{P} \sum_{k=0}^{P}  \frac{\mathbf{x}_{t+k}^{(i)} - \mu^{(i)}}{\sigma^{(i)}}  \cdot 
     \frac{\mathbf{x}_{t'+k}^{(j)} - \mu^{(j)}}{\sigma^{(j)}} 
\end{equation}
where $P=\{32,64,96,128\}$ is the patch size to measure correlations, $\mu^{(\cdot)}$ and $\sigma^{(\cdot)}$ are the mean and standard deviation of corresponding time series patches.

\begin{figure}[htbp]
    \centering
    \begin{subfigure}{\columnwidth} 
        \includegraphics[width=\linewidth]{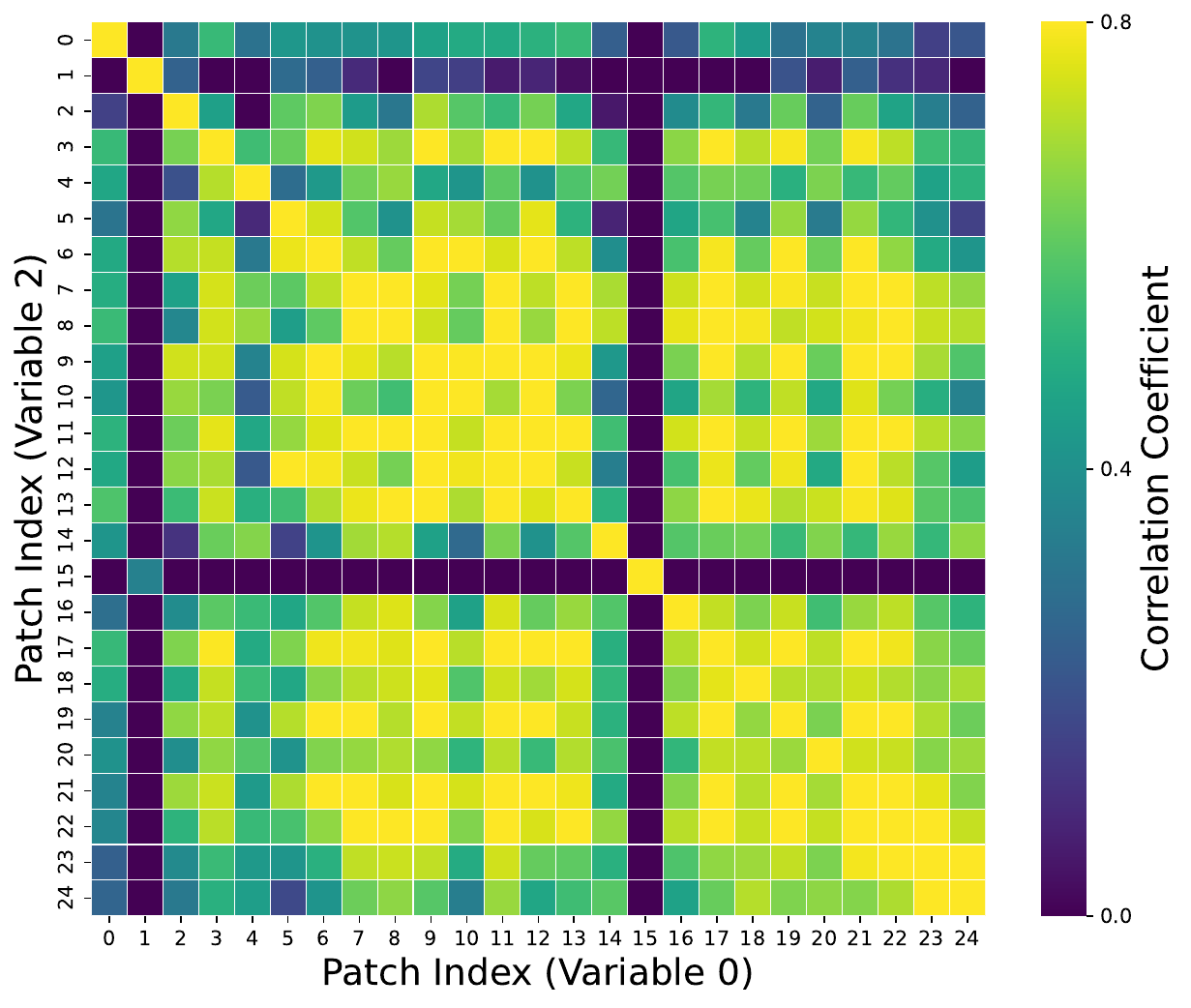}
        \caption{Correlation Matrix: ETTm2 Dataset}
        \label{fig:sub1}
    \end{subfigure}
    \vfill 
    \begin{subfigure}{\columnwidth} 
        \includegraphics[width=\linewidth]{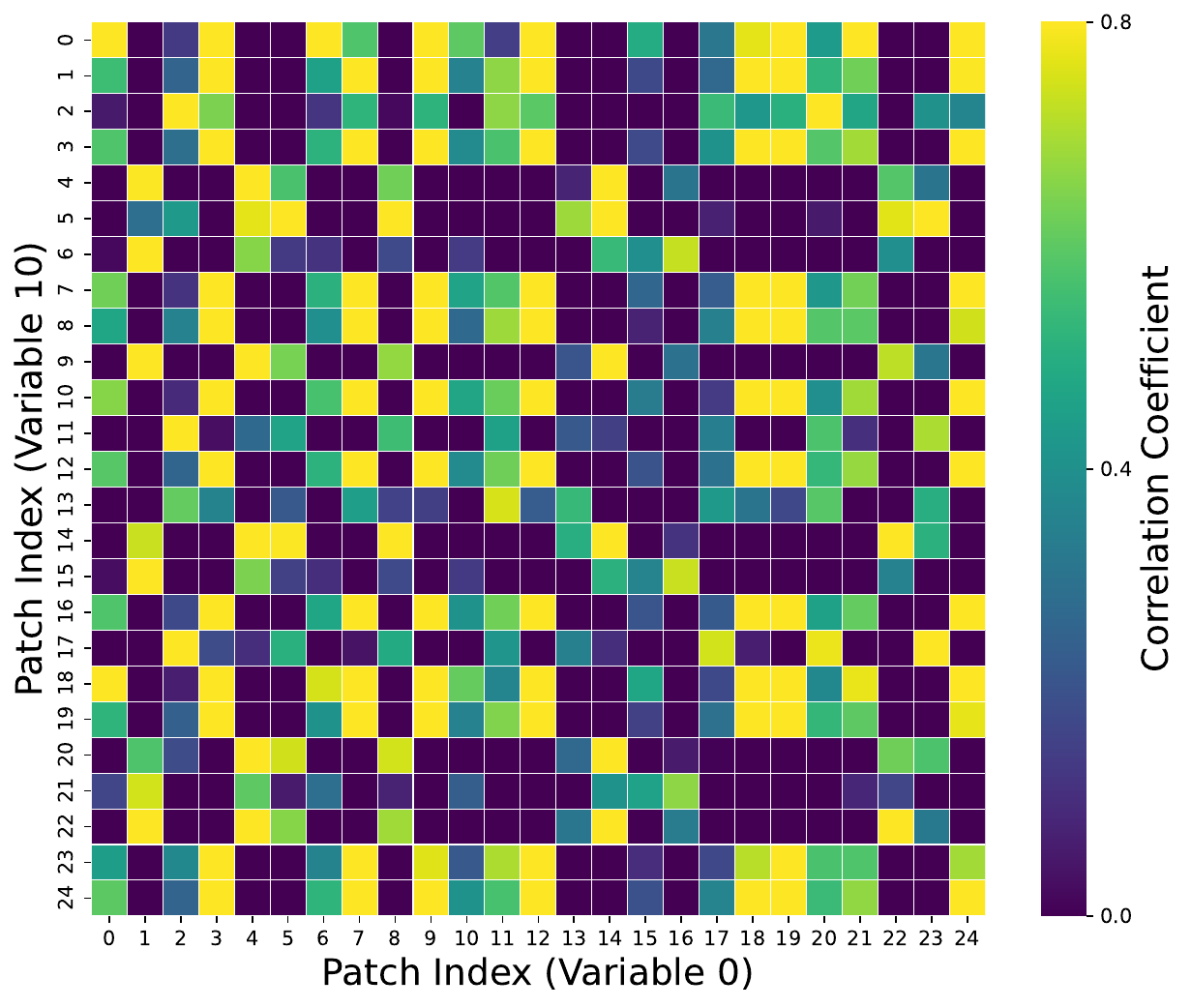}
        \caption{Correlation Matrix: Weather Dataset}
        \label{fig:sub2}
    \end{subfigure}
    \caption{Heatmap for Cross-Time Cross-Variate Correlation Coefficient of ETTm2 and Weather. The color intensity (brightness) within the heatmaps directly maps to the absolute magnitude of the Cross-Time Cross-Variate Correlation Coefficient. Bright regions (represented by lighter yellow) indicate \enquote{correlation hotspots} where the coefficient approaches 0.8, signifying potent synchronous or asynchronous dependencies between the specific variable patches. Conversely, dark regions (represented by deep purple) denote \enquote{null zones} where the correlation is near zero, indicating that the variables evolve independently without identifiable inter-series relationships in those temporal windows.}
    \label{fig:cv}
\end{figure}

Table \ref{tab:correlation_results} presents averaged results under different patch sizes across five benchmark datasets. We use several patch lengths to ensure measurement reliability. Notably, ETT series datasets exhibit significantly stronger correlations. In contrast, the Weather dataset shows consistently weak correlations ($<0.0011$) across all patch sizes.
Figure~\ref{fig:cv} visualizes the \textbf{Cross-Time Cross-Variate Correlation Coefficients} for the first 25 patches (patch size=96) of two variables selected from the public datasets ETTm2 and Weather.

\begin{table}[ht]  
\centering  
\caption{Cross-Time Cross-Variate Correlation Coefficients}  
\label{tab:correlation_results}  
\begin{tabular}{c|cccc|c}  
\toprule  
\multirow{2}{*}{\textbf{Dataset}} & \multicolumn{5}{c}{\textbf{Patch Size } \textit{P}} \\  
\cmidrule(lr){2-6}  
& 32 & 64 & 96 & 128 & Avg\\  
\midrule  
ETTh1 & 0.0134 & 0.0116 & 0.0598 & 0.0105 & 0.0238\\  
ETTh2 & 0.0087& 0.0080	&0.1094&	0.0075&	0.0334\\  
ETTm1 & 0.0306&	0.0179	&0.0624	&0.0148&	0.0314\\  
ETTm2 & 0.0143&	0.0101&	0.1347&	0.0090&	0.0420\\  
Weather & 0.0007	&0.0011&	0.0009&	0.0002&	\textbf{0.0007}\\  
\bottomrule  
\end{tabular}  
\end{table}

In Section \ref{Robustness to Irrelevant Dimensions}, we evaluate the robustness of DualWeaver against irrelevant dimensions. By injecting non-informative noise sequences into the ETTh1 dataset, we simulate scenarios where cross-variable correlations progressively weaken. Empirical results demonstrate that while the forecasting performance of AdaPTS and Chronos-2 degrades as the number of noise channels increases, DualWeaver maintains a remarkably stable error profile, showcasing its superior resilience to spurious features. Table \ref{tab:noise_correlation_results} provides detailed Cross-Time Cross-Variate Correlation Coefficient measurements for ETTh1 under increasing noise injection.

\begin{table}[ht]  
\centering  
\caption{Cross-variable correlation decay under synthetic noise injection.}  
\label{tab:noise_correlation_results}  
\begin{tabular}{c|cccc|c}  
\toprule  
\textbf{Noise} & \multicolumn{5}{c}{\textbf{Patch Size } \textit{P}} \\  
\cmidrule(lr){2-6}
\textbf{Channels} & 32 & 64 & 96 & 128 & Avg\\  
\midrule  
0 & 0.0134	& 0.0116	& 0.0598	& 0.0105	& 0.0238 \\  
1 & 0.0105	& 0.0087	& 0.0461	& 0.0080	& 0.0183 \\
2 & 0.0078	& 0.0073	& 0.0364	& 0.0065	& 0.0145 \\
3 & 0.0068	& 0.0057	& 0.0294	& 0.0052	& 0.0118 \\
4 & 0.0055	& 0.0049	& 0.0246	& 0.0044	& 0.0098 \\
5 & 0.0046	& 0.0039	& 0.0199	& 0.0038	& 0.0081 \\
6 & 0.0039	& 0.0037	& 0.0174	& 0.0033	& 0.0071 \\
7 & 0.0033	& 0.0030	& 0.0145	& 0.0028	& 0.0059 \\
8 & 0.0027	& 0.0026	& 0.0131	& 0.0024	& 0.0052 \\
9 & 0.0027	& 0.0022	& 0.0113	& 0.0022	& 0.0046 \\
10 & 0.0023	& 0.0022	& 0.0098	& 0.0023	& 0.0042 \\

\bottomrule  
\end{tabular}  
\end{table}

\section{Training Stability From Dual-Surrogate}\label{Training Stability From Dual-Surrogate}
The dual-surrogate mechanism in DualWeaver bolsters training stability by leveraging complementary optimization directions. To isolate this effect, we introduce SingleWeaver, a baseline formulated with only a single surrogate: $\mathbf{S}_\text{single}  = f(\mathbf X)+ \mathbf{w_\alpha} \odot \mathbf X$. This ablation baseline enables a direct comparison of optimization stability between single-surrogate and dual-surrogate strategies.

\begin{figure}[htbp]
  \centering
  \includegraphics[width=0.9\columnwidth]{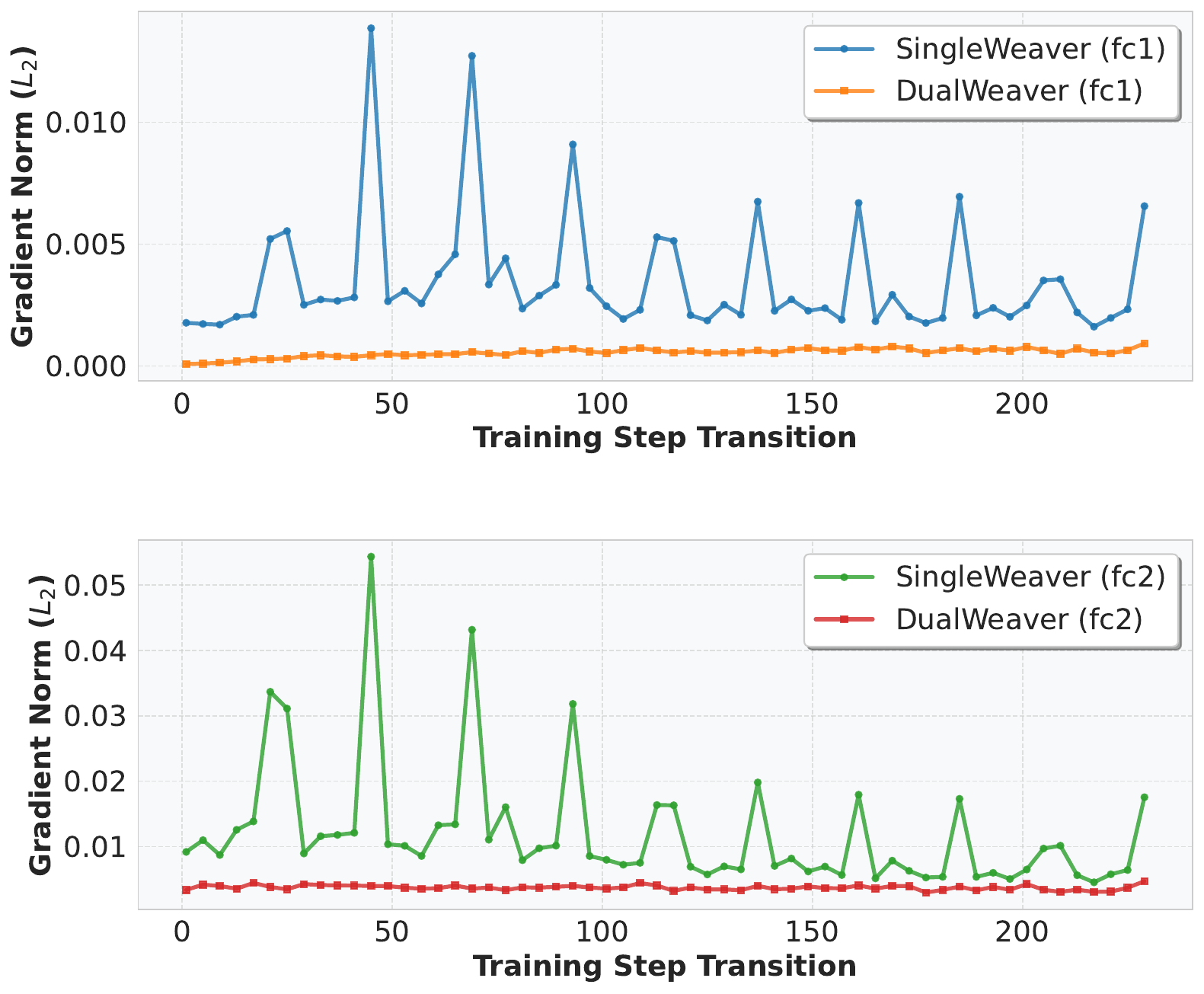}
  \caption{Comparison of gradient dynamics and training stability. We visualize the $L_{2}$ norm of weight gradient differences between consecutive steps ($\nabla \mathbf{W}_{t+1} - \nabla \mathbf{W}_t$) for the fully-connected layers of the feature-fusion module. DualWeaver maintains a remarkably smooth, consistent gradient transition, underscoring the superior numerical stability of our dual-surrogate optimization framework.}
  \label{fig:gradient}
\end{figure}

We monitor the stability of gradient updates in the fully connected layers (fc1 and fc2) of the feature-fusion module. Specifically, we compute the Euclidean distance (i.e., the $L_{2}$ norm of the difference) between the weight gradients of consecutive training steps ($t$ and $t+1$) to characterize the magnitude of instantaneous gradient fluctuations. Figure \ref{fig:gradient} presents the visualized results, with values aggregated every four steps to enhance clarity. The results demonstrate that while DualWeaver maintains a remarkably smooth gradient transition, the SingleWeaver suffers from frequent, high-magnitude spikes. This contrast underscores that our dual-tuned approach effectively mitigates optimization instability during adaptation.

\section{Full Results}

Comprehensive experimental results across the primary public benchmark datasets are provided in Table \ref{detail public dataset results}. The \textbf{\textcolor{red}{bold}} and \textcolor{blue}{\underline{underlined}} respectively denote the best and second-best results for each horizon. Experimental results for TimePro, SimpleTM, TimeMixer, and iTransformer are officially reported in \cite{TimePro} and \cite{simpletm}.

\begin{table*}[ht]
\centering
\caption{Multivariate forecasting results on public datasets with forecasting horizons $\in \{96, 192, 336, 720\}$.}
\begin{center}

\resizebox{\linewidth}{!}{
\begin{tabular}{c|c|c|cc|cc|cc|cc|cc|cc|cc|cc}
\toprule

\multicolumn{3}{c|}{\multirow{2}{*}{\textbf{Methods}}} & 
\multicolumn{2}{c}{\textbf{DualWeaver}} & 
\multicolumn{2}{c}{\textbf{AdaPTS}} & 
\multicolumn{2}{c}{\multirow{2}{*}{\textbf{SFT}}} & 
\multicolumn{2}{c}{\multirow{2}{*}{\textbf{Vanilla}}} & 
\multicolumn{2}{c}{\textbf{TimePro}} & 
\multicolumn{2}{c}{\textbf{SimpleTM}} & 
\multicolumn{2}{c}{\textbf{TimeMixer}} & 
\multicolumn{2}{c}{\textbf{iTrans.}} \\
\multicolumn{3}{c|}{} & 
\multicolumn{2}{c}{\textbf{(Ours)}} & 
\multicolumn{2}{c}{(2025)} & 
\multicolumn{2}{c}{} &  
\multicolumn{2}{c}{} &  
\multicolumn{2}{c}{(2025)} & 
\multicolumn{2}{c}{(2025)} & 
\multicolumn{2}{c}{(2024)} & 
\multicolumn{2}{c}{(2024)} \\ 

\cmidrule(lr){1-3} \cmidrule(lr){4-5} \cmidrule(lr){6-7} \cmidrule(lr){8-9} \cmidrule(lr){10-11}
\cmidrule(lr){12-13} \cmidrule(lr){14-15} \cmidrule(lr){16-17} \cmidrule(lr){18-19}

\multicolumn{3}{c|}{\textbf{Metirc}} & MSE & MAE & MSE & MAE & MSE & MAE & MSE & MAE & MSE & MAE & MSE & MAE & MSE & MAE & MSE & MAE \\

\midrule[\heavyrulewidth]


& \multirow{5}{*}{\rotatebox{90}{ETTh1}}
& 96
& \textcolor{blue}{\underline{0.352}}	& \textbf{\textcolor{red}{0.380}}	& \textbf{\textcolor{red}{0.351}}	& 0.387
& 0.360	& \textcolor{blue}{\underline{0.384}}
& 0.366 & 0.392 
& 0.375 & 0.398 & 0.366 & 0.392 & 0.381 & 0.401 & 0.386 & 0.405 
\\

&& 192
& \textcolor{blue}{\underline{0.401}} &	\textbf{\textcolor{red}{0.410}} &	\textbf{\textcolor{red}{0.401}} &	0.421	
& 0.411	& \textcolor{blue}{\underline{0.415}}
& 0.415 & 0.419 
& 0.427 & 0.429 & 0.422 & 0.421 & 0.440 & 0.433 & 0.441 & 0.436 
\\

&& 336
& \textbf{\textcolor{red}{0.434}}	& \textbf{\textcolor{red}{0.431}}	& \textcolor{blue}{\underline{0.437}} &	0.445	
& 0.452	& \textcolor{blue}{\underline{0.438}}
& 0.450 & 0.439 
& 0.472 & 0.450 & 0.440 & 0.438 & 0.501 & 0.462 & 0.487 & 0.458 
\\

&& 720
& \textbf{\textcolor{red}{0.437}} &	\textbf{\textcolor{red}{0.456}}	& 0.456	& 0.479	
& 0.486	& 0.474
& \textcolor{blue}{\underline{0.446}} & \textcolor{blue}{\underline{0.457}} 
& 0.476 & 0.474 & 0.463 & 0.462 & 0.501 & 0.482 & 0.503 & 0.491 
\\
\cmidrule(lr){3-19}

&& Avg
& \textbf{\textcolor{red}{0.406}}	& \textbf{\textcolor{red}{0.419}}	& \textcolor{blue}{\underline{0.411}}	& 0.433	
& 0.427	& 0.428
& 0.419 & \textcolor{blue}{\underline{0.427}}
& 0.438 & 0.438 & 0.422 & 0.428 & 0.458 & 0.445 & 0.454 & 0.447
\\

\cmidrule(lr){2-19}

& \multirow{5}{*}{\rotatebox{90}{ETTh2}}
& 96
& \textbf{\textcolor{red}{0.274}}	& \textbf{\textcolor{red}{0.336}}	& \textcolor{blue}{\underline{0.280}}	& 0.349	
& 0.280	& 0.338
& 0.287 & 0.345 
& 0.293 & 0.345 & 0.281 & \textcolor{blue}{\underline{0.338}} & 0.292 & 0.343 & 0.297 & 0.349 
\\

&& 192
& \textbf{\textcolor{red}{0.334}}	& \textbf{\textcolor{red}{0.378}}	& 0.341	& 0.389	
& \textcolor{blue}{\underline{0.339}} & \textcolor{blue}{\underline{0.378}}
& 0.347 & 0.385 
& 0.367 & 0.394 & 0.355 & 0.387 & 0.374 & 0.395 & 0.380 & 0.400 
\\

&& 336
& \textcolor{blue}{\underline{0.371}}	& \textcolor{blue}{\underline{0.404}}	& 0.376	& 0.416	
& 0.377	& 0.406
& 0.387 & 0.412 
& 0.419 & 0.431 & \textbf{\textcolor{red}{0.365}} & \textbf{\textcolor{red}{0.401}} & 0.428 & 0.433 & 0.428 & 0.432 
\\

&& 720
& \textbf{\textcolor{red}{0.383}}	& \textbf{\textcolor{red}{0.426}}	& \textcolor{blue}{\underline{0.395}}	& 0.440	
& 0.401	& \textcolor{blue}{\underline{0.433}}
& 0.413 & 0.440 
& 0.427 & 0.445 & 0.413 & 0.436 & 0.454 & 0.458 & 0.427 & 0.445
\\
\cmidrule(lr){3-19}

&& Avg
& \textbf{\textcolor{red}{0.340}}	& \textbf{\textcolor{red}{0.386}}	& \textcolor{blue}{\underline{0.348}}	& 0.399	
& 0.349	& \textcolor{blue}{\underline{0.389}}
& 0.359 & 0.396 
& 0.377 & 0.403 & 0.353 & 0.391 & 0.384 & 0.407 & 0.383 & 0.407
\\

\cmidrule(lr){2-19}

\multirow{8}{*}{\textbf{\rotatebox{90}{TimerXL}}} 

& \multirow{5}{*}{\rotatebox{90}{ETTm1}}
& 96 
& \textbf{\textcolor{red}{0.272}}	& \textbf{\textcolor{red}{0.330}}	& \textcolor{blue}{\underline{0.273}}	& 0.338	
& 0.285	& \textcolor{blue}{\underline{0.334}}
& 0.307 & 0.355 
& 0.326 & 0.364 & 0.321 & 0.361 & 0.328 & 0.363 & 0.334 & 0.368
\\

&& 192
& \textbf{\textcolor{red}{0.306}}	& \textbf{\textcolor{red}{0.356}}	& \textcolor{blue}{\underline{0.318}}	& 0.371	
& 0.331	& \textcolor{blue}{\underline{0.363}}
& 0.356 & 0.386 
& 0.367 & 0.383 & 0.360 & 0.380 & 0.364 & 0.384 & 0.377 & 0.391
\\

&& 336
& \textbf{\textcolor{red}{0.343}}	& \textbf{\textcolor{red}{0.386}}	& \textcolor{blue}{\underline{0.355}}	& 0.401	
& 0.384	& \textcolor{blue}{\underline{0.398}}
& 0.411 & 0.419 
& 0.402 & 0.409 & 0.390 & 0.404 & 0.390 & 0.404 & 0.426 & 0.420
\\

&& 720
& \textbf{\textcolor{red}{0.430}}	& \textcolor{blue}{\underline{0.444}} &	\textcolor{blue}{\underline{0.442}}	& 0.460	
& 0.507	& 0.466
& 0.514 & 0.477 
& 0.469 & 0.446 & 0.454 & \textbf{\textcolor{red}{0.438}} & 0.458 & 0.445 & 0.491 & 0.459
\\
\cmidrule(lr){3-19}

&& Avg
& \textbf{\textcolor{red}{0.338}}	& \textbf{\textcolor{red}{0.379}}	& \textcolor{blue}{\underline{0.347}}	& 0.392	
& 0.377	& \textcolor{blue}{\underline{0.390}}
& 0.397 & 0.409 
& 0.391 & 0.400 & 0.381 & 0.396 & 0.385 & 0.399 & 0.407 & 0.410 
\\

\cmidrule(lr){2-19}

& \multirow{5}{*}{\rotatebox{90}{ETTm2}}
& 96
& \textbf{\textcolor{red}{0.158}}	& \textbf{\textcolor{red}{0.246}}	& 0.166	& 0.268	
& \textcolor{blue}{\underline{0.161}}	& \textcolor{blue}{\underline{0.248}}
& 0.191 & 0.279 
& 0.178 & 0.260 & 0.173 & 0.257 & 0.176 & 0.259 & 0.180 & 0.264
\\

&& 192
& \textcolor{blue}{\underline{0.219}}	& \textbf{\textcolor{red}{0.291}}	& 0.228	& 0.311	
& \textbf{\textcolor{red}{0.219}}	& \textcolor{blue}{\underline{0.292}}
& 0.244 & 0.317 
& 0.242 & 0.303 & 0.238 & 0.299 & 0.242 & 0.303 & 0.250 & 0.309
\\

&& 336
& \textcolor{blue}{\underline{0.281}}	& \textbf{\textcolor{red}{0.335}}	& 0.290	& 0.352	
& \textbf{\textcolor{red}{0.280}}	& \textcolor{blue}{\underline{0.336}}
& 0.299 & 0.357 
& 0.303 & 0.342 & 0.296 & 0.338 & 0.304 & 0.342 & 0.311 & 0.348
\\

&& 720
& \textbf{\textcolor{red}{0.382}}	& \textcolor{blue}{\underline{0.398}}	& 0.389	& 0.414	
& \textcolor{blue}{\underline{0.385}}	& 0.403
& 0.416 & 0.434 
& 0.400 & 0.399 & 0.393 & \textbf{\textcolor{red}{0.395}} & 0.393 & 0.397 & 0.412 & 0.407
\\

\cmidrule(lr){3-19}

&& Avg
& \textbf{\textcolor{red}{0.260}}	& \textbf{\textcolor{red}{0.318}}	& 0.268	& 0.336	
& \textcolor{blue}{\underline{0.261}}	& \textcolor{blue}{\underline{0.320}}
& 0.288 & 0.347 
& 0.281 & 0.326 & 0.275 & 0.322 & 0.278 & 0.325 & 0.288 & 0.332 
\\

\cmidrule(lr){2-19}

& \multirow{5}{*}{\rotatebox{90}{Weather}}
& 96 
& \textbf{\textcolor{red}{0.146}}	& \textbf{\textcolor{red}{0.199}}	& 0.151	& 0.220	
& \textcolor{blue}{\underline{0.149}}	& \textcolor{blue}{\underline{0.204}}
& 0.169 & 0.225 
& 0.166 & 0.207 & 0.162 & 0.207 & 0.165 & 0.212 & 0.174 & 0.214
\\

&& 192
& \textbf{\textcolor{red}{0.192}}	& \textbf{\textcolor{red}{0.246}}	&\textcolor{blue}{\underline{0.194}}	& 0.259	
& 0.194	& 0.250
& 0.212 & 0.267 
& 0.216 & 0.254 & 0.208 & \textcolor{blue}{\underline{0.248}} & 0.209 & 0.253 & 0.221 & 0.254
\\

&& 336
& \textbf{\textcolor{red}{0.242}}	& \textbf{\textcolor{red}{0.289}}	& 0.244	& 0.308	
& \textcolor{blue}{\underline{0.243}} & 0.292
& 0.270 & 0.318 
& 0.273 & 0.296 & 0.263 & \textcolor{blue}{\underline{0.290}} & 0.264 & \underline{0.293} & 0.278 & 0.296
\\

&& 720
& \textcolor{blue}{\underline{0.322}}	& 0.348	& \textbf{\textcolor{red}{0.318}}	& 0.350	
& 0.326	& 0.354
& 0.367 & 0.393 
& 0.351 & 0.346 & 0.340 & \textbf{\textcolor{red}{0.341}} & 0.342 & \textcolor{blue}{\underline{0.345}} & 0.358 & 0.347
\\

\cmidrule(lr){3-19}

&& Avg
& \textbf{\textcolor{red}{0.225}}	& \textbf{\textcolor{red}{0.271}}	& \textcolor{blue}{\underline{0.227}}	& 0.284	
& 0.228	& 0.275 
& 0.254 & 0.301 
& 0.251 & 0.276 & 0.243 & \textcolor{blue}{\underline{0.271}} & 0.245 & 0.276 & 0.258 & 0.278
\\

\midrule

& \multirow{5}{*}{\rotatebox{90}{ETTh1}}
& 96 
& \textbf{\textcolor{red}{0.340}}	& \textbf{\textcolor{red}{0.379}}	& \textcolor{blue}{\underline{0.344}}	& 0.385	
& 0.349	& \textcolor{blue}{\underline{0.385}}	
& 0.354	& 0.390
& 0.375 & 0.398 & 0.366 & 0.392 & 0.381 & 0.401 & 0.386 & 0.405
\\

&& 192
& \textbf{\textcolor{red}{0.381}}	& \textbf{\textcolor{red}{0.406}}	& \textcolor{blue}{\underline{0.384}}	& \textcolor{blue}{\underline{0.410}}
&	0.391	& 0.413	
& 0.400	& 0.423
& 0.427 & 0.429 & 0.422 & 0.421 & 0.440 & 0.433 & 0.441 & 0.436 
\\

&& 336
& \textbf{\textcolor{red}{0.405}}	& \textbf{\textcolor{red}{0.422}}	& 0.433	& 0.445	
& \textcolor{blue}{\underline{0.420}}	& \textcolor{blue}{\underline{0.433}}	
& 0.430	& 0.445
& 0.472 & 0.450 & 0.440 & 0.438 & 0.501 & 0.462 & 0.487 & 0.458 
\\

&& 720
& \textbf{\textcolor{red}{0.455}}	& \textcolor{blue}{\underline{0.470}}	& 0.487	& 0.494	
& 0.484	& 0.493	
& 0.493	& 0.499
& 0.476 & 0.474 & \textcolor{blue}{\underline{0.463}} & \textbf{\textcolor{red}{0.462}} & 0.501 & 0.482 & 0.503 & 0.491 
\\

\cmidrule(lr){3-19}

&& Avg 
& \textbf{\textcolor{red}{0.395}}	& \textbf{\textcolor{red}{0.419}}	& 0.412	& 0.434	
& \textcolor{blue}{\underline{0.411}}	& 0.431
& 0.419	& 0.439
& 0.438 & 0.438 & 0.422 & \textcolor{blue}{\underline{0.428}} & 0.458 & 0.445 & 0.454 & 0.447
\\

\cmidrule(lr){2-19}

& \multirow{5}{*}{\rotatebox{90}{ETTh2}}
& 96 
& \textbf{\textcolor{red}{0.268}}	& \textbf{\textcolor{red}{0.330}}	& 0.273	 & 0.340	
& \textcolor{blue}{\underline{0.271}}	& \textcolor{blue}{\underline{0.331}}
& 0.274	& 0.335
& 0.293 & 0.345 & 0.281 & 0.338 & 0.292 & 0.343 & 0.297 & 0.349 
\\

&& 192
& \textbf{\textcolor{red}{0.325}}	& \textbf{\textcolor{red}{0.375}}	& 0.334	& 0.385	
& \textcolor{blue}{\underline{0.329}}	& \textcolor{blue}{\underline{0.377}}	
& 0.331	& 0.379
& 0.367 & 0.394 & 0.355 & 0.387 & 0.374 & 0.395 & 0.380 & 0.400
\\

&& 336
& \textbf{\textcolor{red}{0.351}}	& \textbf{\textcolor{red}{0.399}}	& 0.361	& 0.408	
& \textcolor{blue}{\underline{0.358}}	& 0.403	
& 0.358	& 0.404
& 0.419 & 0.431 & 0.365 & \textcolor{blue}{\underline{0.401}} & 0.428 & 0.433 & 0.428 & 0.432 
\\
&& 720
& 0.388	& \textbf{\textcolor{red}{0.436}}	& 0.421	& 0.458	
& \textbf{\textcolor{red}{0.387}}	& 0.437	
& \textcolor{blue}{\underline{0.387}}	& 0.438
& 0.427 & 0.445 & 0.413 & \textcolor{blue}{\underline{0.436}} & 0.454 & 0.458 & 0.427 & 0.445 
\\
\cmidrule(lr){3-19}

&& Avg
& \textbf{\textcolor{red}{0.333}}	& \textbf{\textcolor{red}{0.385}}	& 0.347	& 0.398	
& \textcolor{blue}{\underline{0.336}}	& \textcolor{blue}{\underline{0.387}}
& 0.337	& 0.389
& 0.377 & 0.403 & 0.353 & 0.391 & 0.384 & 0.407 & 0.383 & 0.407 
\\

\cmidrule(lr){2-19}

\multirow{5}{*}{\textbf{\rotatebox{90}{Sundial}}} 
& \multirow{5}{*}{\rotatebox{90}{ETTm1}}
& 96 
& \textbf{\textcolor{red}{0.261}}	& \textbf{\textcolor{red}{0.324}}	& \textcolor{blue}{\underline{0.268}}	& 0.334	
& 0.288	& \textcolor{blue}{\underline{0.333}}	
& 0.285	 & 0.337
& 0.326 & 0.364 & 0.321 & 0.361 & 0.328 & 0.363 & 0.334 & 0.368
\\

&& 192
& \textbf{\textcolor{red}{0.307}}	& \textbf{\textcolor{red}{0.355}}	& \textcolor{blue}{\underline{0.313}}	& \textcolor{blue}{\underline{0.365}}
& 0.334	& 0.366	
& 0.325	& 0.369
& 0.367 & 0.383 & 0.360 & 0.380 & 0.364 & 0.384 & 0.377 & 0.391
\\

&& 336
& \textbf{\textcolor{red}{0.333}}	& \textbf{\textcolor{red}{0.374}}	& \textcolor{blue}{\underline{0.342}}	& 0.387	
& 0.354	& \textcolor{blue}{\underline{0.385}}	
& 0.353	& 0.391
& 0.402 & 0.409 & 0.390 & 0.404 & 0.390 & 0.404 & 0.426 & 0.420
\\

&& 720
& \textbf{\textcolor{red}{0.375}}	& \textbf{\textcolor{red}{0.403}}	& 0.393	& 0.424	
& \textcolor{blue}{\underline{0.387}}	& \textcolor{blue}{\underline{0.411}}	
& 0.395	& 0.420
& 0.469 & 0.446 & 0.454 & 0.438 & 0.458 & 0.445 & 0.491 & 0.459 
\\
\cmidrule(lr){3-19}

&& Avg
& \textbf{\textcolor{red}{0.319}}	& \textbf{\textcolor{red}{0.364}}	& \textcolor{blue}{\underline{0.329}}	& 0.377	
& 0.341	& \textcolor{blue}{\underline{0.374}}	
& 0.340	& 0.379
& 0.391 & 0.400 & 0.381 & 0.396 & 0.385 & 0.399 & 0.407 & 0.410 
\\

\cmidrule(lr){2-19}

& \multirow{5}{*}{\rotatebox{90}{ETTm2}}

& 96 
& \textbf{\textcolor{red}{0.153}}	& \textbf{\textcolor{red}{0.240}}	& 0.163	& 0.267	
& \textcolor{blue}{\underline{0.155}}	& \textcolor{blue}{\underline{0.240}}	
& 0.173 & 0.259
& 0.178 & 0.260 & 0.173 & 0.257 & 0.176 & 0.259 & 0.180 & 0.264
\\

&& 192
& \textbf{\textcolor{red}{0.207}}	& \textbf{\textcolor{red}{0.282}}	& 0.231	& 0.325	
& \textcolor{blue}{\underline{0.212}}	& \textcolor{blue}{\underline{0.283}}	
& 0.231	& 0.303
& 0.242 & 0.303 & 0.238 & 0.299 & 0.242 & 0.303 & 0.250 & 0.309
\\
&& 336
& \textbf{\textcolor{red}{0.259}}	& \textbf{\textcolor{red}{0.320}}	& 0.294	& 0.371	
& \textcolor{blue}{\underline{0.265}}	& \textcolor{blue}{\underline{0.320}}	
& 0.284	& 0.340
& 0.303 & 0.342 & 0.296 & 0.338 & 0.304 & 0.342 & 0.311 & 0.348 
\\
&& 720
& \textbf{\textcolor{red}{0.336}}	& \textcolor{blue}{\underline{0.373}}	& 0.383	& 0.426	
& \textcolor{blue}{\underline{0.341}}	& \textbf{\textcolor{red}{0.372}}
& 0.355	& 0.391
& 0.400 & 0.399 & 0.393 & 0.395 & 0.393 & 0.397 & 0.412 & 0.407 
\\

\cmidrule(lr){3-19}

&& Avg
& \textbf{\textcolor{red}{0.239}}	& \textbf{\textcolor{red}{0.304}}	& 0.268	& 0.347	
& \textcolor{blue}{\underline{0.243}}	& \textcolor{blue}{\underline{0.304}}	
& 0.261	& 0.323
& 0.281 & 0.326 & 0.275 & 0.322 & 0.278 & 0.325 & 0.288 & 0.332 
\\

\cmidrule(lr){2-19}
 
& \multirow{5}{*}{\rotatebox{90}{Weather}}
& 96 
& \textbf{\textcolor{red}{0.137}}	& \textbf{\textcolor{red}{0.187}}	& 0.139	& 0.206	
& \textcolor{blue}{\underline{0.138}}	& \textcolor{blue}{\underline{0.187}}	
& 0.160	& 0.208
& 0.166 & 0.207 & 0.162 & 0.207 & 0.165 & 0.212 & 0.174 & 0.214 
\\

&& 192
& \textbf{\textcolor{red}{0.180}}	& \textbf{\textcolor{red}{0.233}}	& 0.184	& 0.254	
& \textcolor{blue}{\underline{0.181}}	& \textcolor{blue}{\underline{0.233}}	
& 0.209	& 0.255
& 0.216 & 0.254 & 0.208 & 0.248 & 0.209 & 0.253 & 0.221 & 0.254 
\\
&& 336
& \textbf{\textcolor{red}{0.228}}	& \textcolor{blue}{\underline{0.275}}	& 0.233	& 0.297	
& \textcolor{blue}{\underline{0.229}}	& \textbf{\textcolor{red}{0.273}}	
& 0.258	& 0.292
& 0.273 & 0.296 & 0.263 & 0.290 & 0.264 & 0.293 & 0.278 & 0.296 
\\
&& 720
& \textbf{\textcolor{red}{0.292}}	& \textcolor{blue}{\underline{0.323}}	& \textcolor{blue}{\underline{0.292}}	& 0.342	
& 0.293	& \textbf{\textcolor{red}{0.322}}
& 0.326	& 0.339
& 0.351 & 0.346 & 0.340 & 0.341 & 0.342 & 0.345 & 0.358 & 0.347 
\\

\cmidrule(lr){3-19}

&& Avg
& \textbf{\textcolor{red}{0.209}}	& \textcolor{blue}{\underline{0.255}}	& 0.212	& 0.275	
& \textcolor{blue}{\underline{0.210}}	& \textbf{\textcolor{red}{0.254}}	
& 0.238	& 0.274
& 0.251 & 0.276 & 0.243 & 0.271 & 0.245 & 0.276 & 0.258 & 0.278 
\\

\midrule
\multicolumn{3}{c|}{\textbf{$1^{st}$}}
& \textcolor{red}{\textbf{33}} & \textcolor{red}{\textbf{32}} & \textcolor{blue}{\underline{3}} & 0 
& \textcolor{blue}{\underline{3}} & 3
& 0 & 0 
& 0 & 0 & 1 & \textcolor{blue}{\underline{5}} & 0 & 0 &	0 & 0	
\\

\bottomrule
\end{tabular}

}

\label{detail public dataset results}
\end{center}
\end{table*}

\end{document}